\documentclass[extra,mreferee]{gji}
\usepackage{timet}
\usepackage{amsmath,amssymb,bm}
\usepackage{graphicx}
\usepackage{subfig}
\usepackage{booktabs} 
\allowdisplaybreaks[4]

\title[Bayesian geoacoustic inversion using MDN]{Bayesian geoacoustic inversion using mixture density network}

\author[Wu et al]
  {Guoli Wu$^{1}$, Jingya Zhang$^{2}$ and  Junqiang Song$^3$\\
  $^1$ Intelligent Game and Decision Lab, Beijing, China. E-mail: wuguolihang94@163.com \\
  $^2$ Research Institute of Petroleum Exploration \& Development, PetroChina, Beijing,  China \\
  $^3$ College of Meteorology and Oceanography, National University of Defense Technology, Changsha, China
  }
  
\date{Received 2020 July 18; in original form 2020 July 18}
\pagerange{\pageref{firstpage}--\pageref{lastpage}}

\newcommand{\q}[1]{``#1''}
\begin{document}

\label{firstpage}

\maketitle

\begin{summary}
 Bayesian geoacoustic inversion problems are conventionally solved by Markov chain Monte Carlo methods or its variants, which are computationally expensive. This paper extends the classic Bayesian geoacoustic inversion framework by deriving important geoacoustic statistics of Bayesian geoacoustic inversion from the multidimensional posterior probability density (PPD) using the mixture density network (MDN) theory. These statistics make it convenient to train the network directly on the whole parameter space and get the multidimensional PPD of model parameters. The present approach provides a much more efficient way to solve geoacoustic inversion problems in Bayesian inference framework. The network is trained on a simulated dataset of surface-wave dispersion curves with shear-wave velocities as labels and tested on both synthetic and real data cases. The results show that the network gives reliable predictions and has good generalization performance on unseen data. Once trained, the network can rapidly (within seconds) give a fully probabilistic solution which is comparable to Monte Carlo methods. It provides an promising approach for real-time inversion.
\end{summary}


\section{Introduction}


Geoacoustic inversion focuses on the reconstruction of unknown geoacoustic model parameters (such as wave speed, density, attenuation and layer thickness, etc.), which can not be directly observed, from a set of observations. It is a challenging and representative inverse problem because of its highly nonlinearity and non-uniqueness and has drawn a great deal of attention in seismic and underwater acoustics communities. Some researchers use linearized inversions to approximate the nonlinear relation between the model and the observations by a linearized relation \citep{Tarantola1982,Caiti1994,zhdanov2002,Bensen2009,hefengdong2010}. They apply regularization techniques to deal with the ill-conditioning problem and seek an approximate solution which can minimize the misfit using optimization algorithms. Some other researchers use metaheuristic algorithms, such as genetic algorithms \citep{Gerstoft1994,Heard1998}, simulated annealing algorithms \citep{Collins1992,Dosso1993,Gerstoft2000} and hybrid inversion algorithms \citep{Gerstoft1995,Musil1999,Dosso2001}, to search the model space and approximate the global optimal solutions. These nonlinear optimization methods provide better capability in searching the global optimization model that fits the observations best than linearized inversions. However, the lack of quantitative uncertainty estimates limits the understanding of the solutions especially when the solution of the data is non-unique.

Bayesian inference provides a probabilistic framework to quantify parameter uncertainties for inversion problems. In a Bayesian formulation, the general solution to an inversion problem is expressed as the posterior probability density (PPD) of the model parameters given the observed data. The model parameter estimates and uncertainties can be characterized by the moments of the multidimensional PPD, such as the optimal model estimates (e.g., the posterior mean model, the maximum a \emph{posteriori} (MAP) model), the marginal probability density function (PDF) of the model parameter and the inter-parameter relationships (covariance matrix). Bayesian inversion has been applied to solve geoacoustic inverse problem since the 1990s \citep{Gerstoft1994, Mosegaard1995} and widely used during the past two decades \citep[e.g., ][]{Dosso2002,Dosso2006,Bodin2009,Dong2011,Dosso2011,Steininger2014,Galetti2017,Zhang2018a}. In Bayesian inversion, how to efficiently obtain the asymptotically unbiased samples from the PPD is a key issue. In most publications, sampling based methods, such as Markov-chain Monte Carlo algorithms and its variants, are normally used to sample from the PPD. However, the sampling based methods are computationally intensive and the computational cost is extremely high.

The mixture density network (MDN) is a type of neural network which combines conventional neural networks with a mixture density model \citep{bishop1994mixture}. Different from conventional neural networks, the MDN provides full statistical properties of the solution and is suitable to solve problems involving the prediction of continuous variables such as inverse problems. The MDN was developed by \citet{bishop1994mixture} and applied to a surface wave inversion problem to model the PPD of Moho depth by \citet{Meier2007a}. Later, \citet{Meier2007b} applied the MDN to estimate the 1D marginal PDFs of global crustal parameters. Thereafter, the MDN was applied to nonlinear inverse problems in different fields. \citet{Shahraeeni2012} extended the MDN to perform petrophysical inversion on seismic data. Wit et al. (\citeyear{Wit2013a,Wit2014}) performed Bayesian inference of radial Earth structure using the MDN and obtained 1D marginal PDFs of individual Earth parameters. \citet{Earp2019} applied the MDN to surface wave tomography and presented 1D marginal PDF of subsurface shear-wave velocity for each layer. \citet{Cao2020} applied a similar approach to \citet{Earp2019} but using ambient seismic noise recorded on land. Then the obtained 1D marginal PDFs are used to reconstructed a 3D seismic velocity model. 

These inversion approaches using the MDN are performed under the framework of Bayesian inference. The difference between the traditional Bayesian inversions using sampling based methods and the inversions using the MDN is that sampling based methods try to generate a set of samples from the PPD while the MDN methods directly generate the PPD. Once trained, the MDN normally provides the PPD in seconds, making it a much more time and memory efficient method.

In the previously published applications of the MDN in inverse problems, most researchers performed the MDN on individual model parameters and obtained 1D marginal PDFs \citep{Meier2007a,Meier2007b,Wit2013a,Wit2014,Earp2019,Cao2020}. Only very few researchers addressed the 2D joint PDFs of model parameters \citep{Shahraeeni2012}. In this paper, we propose to treat a set of model parameters as a whole and perform MDN inversion (Bayesian inversion using the MDN) on it, considering that these parameters work together to give an observation in the forward model. In this way we directly take the multidimensional PPD as the output of the MDN and do not need to train individual networks for individual parameters, which would be multifarious when there are many parameters in the model (such as \citet{Earp2019}). Further, we have derived some important geoacoustic statistics based on the multidimensional PPD theoretically and help to avoid time-consuming numerical integration in the application of the MDN inversion. The derivation also improves the understanding of the statistical properties of the Bayesian PPD using the MDN theory. The perspective of the discussion in this paper is limited to geoacoustic inversion, although the derivation and the conclusion is general.

The paper is organized as follows: First, we summarize the basic theory of Bayesian inference and the MDN, and derive the geoacoustic statistics based on the theory. Then we describe the procedure of the MDN inversion, including the preparation of the training set, regularization, network design and training. Further, the evaluation of the network on test set and the results of four synthetic cases and one real data case are presented. Finally, we summarize the work and draw some conclusions.


\section{Theory} 

\subsection{Bayesian inference}

Let $\mathbf{m}$ denotes the $M$-dimensional vector of model parameters with entries $m_i$ and $\mathbf{d}$ denotes the $N$-dimensional vector of the observations with entries $d_i$. According to Bayes' theorem, the posterior probability can be written as
\begin{equation}
      P(\mathbf{m|d}) = \frac{P(\mathbf{d|m})P(\mathbf{m})}{P(\mathbf{d})}.
      \label{pmd}
\end{equation}
$P(\mathbf{m|d})$ denotes the conditional PDF of parameters $\mathbf{m}$ given observations $\mathbf{d}$ and is known as the PPD. $P(\mathbf{m})$ denotes the prior PDF which represents the estimate of the probability of $\mathbf{m}$ independent of observations $\mathbf{d}$. $P(\mathbf{d})$ is the PDF of observations $\mathbf{d}$ and can normally be treated as a constant. $P(\mathbf{d|m})$ denotes conditional PDF of observations $\mathbf{d}$ given parameters $\mathbf{m}$. When the observations $\mathbf{d}$ is fixed, it can be interpreted as the likelihood function and generally be written as 
\begin{equation}
      L(\mathbf{m}) = P(\mathbf{d|m}) \propto {\rm exp}[-E(\mathbf{m})],
      \label{lm}
\end{equation}
where $E(\mathbf{m})$ denotes the misfit function. Some researchers have derived several forms of $E(\mathbf{m})$ under some appropriate assumptions of the distribution of observation errors \citep{Dosso2002,Dosso2011,Dong2011}.
If Equation~\ref{lm} is substituted back into Equation~\ref{pmd}, Equation~\ref{pmd} can be rewritten as
\begin{equation}
      P(\mathbf{m|d}) = kP(\mathbf{m}){\rm exp}[-E(\mathbf{m})],
      \label{pmd1}
\end{equation}
where $k$ is a normalization constant. Taking the logarithm of both sides leads to an expression of the misfit function
\begin{equation}
      E(\mathbf{m}) = -{\rm log}P(\mathbf{m|d})+{\rm log}[kP(\mathbf{m})],
      \label{em}
\end{equation}
where log is the natural logarithm.

The multidimensional PPD is considered as a general solution to Bayesian inversion. Based on the PPD, some important statistics, such as the MAP model, mean model, marginal PDFs  and model covariance matrix \citep{Dosso2011}, can be defined as 
\begin{equation}
      \hat{\mathbf{m}} = \operatorname*{argmax}\limits_{\mathbf{m}} P(\mathbf{m|d}),
      \label{hatm}
\end{equation}
\begin{equation}
      \overline{\mathbf{m}} = \int \mathbf{m} P(\mathbf{m|d}) {\rm d}\mathbf{m},
      \label{overm}
\end{equation}
\begin{equation}
      P(m_i|\mathbf{d}) = \int \delta(m_i-m_i^\prime)  P(\mathbf{m^\prime|d}) {\rm d}\mathbf{m^\prime},
      \label{pmi}
\end{equation}
\begin{equation}
      P(m_i,m_j|\mathbf{d}) = \int \delta(m_i-m_i^\prime)\delta(m_j-m_j^\prime)  P(\mathbf{m^\prime|d}) {\rm d}\mathbf{m^\prime},
      \label{pmimj}
\end{equation}
\begin{equation}
      \mathbf{C}_\mathbf{m} = \int (\mathbf{m}-\overline{\mathbf{m}})(\mathbf{m}-\overline{\mathbf{m}})^T P(\mathbf{m|d}) {\rm d}\mathbf{m},
      \label{cm}
\end{equation}

where $\delta$ denotes the Dirac delta function. Note that $\int$ means integrating over the entire parameter space. Inter-parameter relationships can be computed in a normalized form as 
\begin{equation}
      R_{ij} = C_{m_{ij}}/\sqrt{C_{m_{ii}C_{m_{jj}}}}.
      \label{correlationmatrix}
\end{equation}
Generally speaking, analytical solutions to Equations \ref{hatm}-\ref{cm} are not available for nonlinear inverse problems.

\subsection{Mixture density network}
Assume that we have a data set $D = \{(\mathbf{d}_i,\mathbf{m}_i):i = 1,...,N_D\}$, where $N_D$ is the number of samples, $\mathbf{d}_i$ and $\mathbf{m}_i$ is the observation vector and the corresponding model parameter vector in the $i$th sample, respectively. A neural network can be trained to learn the underlying transformation function between input and output spaces. For the data set $D$, if we take $\mathbf{m}$ as the input and $\mathbf{d}$ as the output, the neural network learns a forward problem. Conversely, it learns an inverse problem. The central goal of the learning is that when fed a new input, the neural network can provide an appropriate prediction as an output. 

For nonlinear inverse problems involving non-unique solutions and prediction of continuous variables, conventional neural networks have limited capability to give adequate statistical information about the solution, while the MDN overcomes this limitation and supply us a framework which has the flexibility to model arbitrary probability distributions \citep{bishop1994mixture,Meier2007a}. The basic idea of the MDN is to model the PPD using a linear combination of Gaussian kernel functions as
\begin{equation}
      P(\mathbf{m|d}) = \sum_{l=1}^{L} \alpha_l(\mathbf{d}) \phi_l(\mathbf{m|d}),
      \label{pmdmdn}
\end{equation}
where $\alpha_l(\mathbf{d})$ denotes the mixing coefficients representing the relative importance of each kernel, $L$ denotes the number of Gaussian kernels used in the mixture and $\phi_l(\mathbf{m|d})$ is the Gaussian kernel function with the form
\begin{equation}
      \phi_l(\mathbf{m|d}) = \frac{1}{(2\pi)^{M/2} \sigma_l(\mathbf{d})^M} {\rm exp} \{-\frac{\left\| \mathbf{m}-\bm{\mu}_l(\mathbf{d})  \right\|^2}{2\sigma_l(\mathbf{d})^2}\}.
      \label{phimd}
\end{equation}
Equation~\ref{phimd} is a multivariate normal distribution of $\mathbf{m}$, in which $\bm{\mu}_l(\mathbf{d})$ is the mean, representing the center of the $l$th kernel and $\sigma_l(\mathbf{d})$ is the common variance. More generally the common variance can be replaced by a full covariance matrix, leading to a more complex form of Equation~\ref{phimd}. However, it is not necessary because such a model with the simplified kernels (Equation~\ref{pmdmdn}) has the potential to approximate any PDF to arbitrary accuracy \citep{mclachlan1988mixture}.

Additionally, according to \citet{bishop1994mixture}, the mixing coefficients are required to satisfy the constraint
\begin{equation}
      \sum_{l=1}^{L} \alpha_l(\mathbf{d}) = 1.
      \label{sumsigma1}
\end{equation}

Note that $\alpha_l(\mathbf{d})$ and $\sigma_l(\mathbf{d})$ are functions of $\mathbf{d}$, respectively, and $\bm{\mu}_l(\mathbf{d})$ is a vector function of $\mathbf{d}$ with the same dimensionality $M$ as $\mathbf{m}$. The $L$ sets of parameters $\alpha_l(\mathbf{d})$, $\sigma_l(\mathbf{d})$ and $\bm{\mu}_l(\mathbf{d})$ fully defined the model and can be used to approximate any possible PPD with a sufficient number of kernels. Therefore, a MDN takes these parameters as its output. Once fed an input $\mathbf{d}_i$, the trained MDN outputs $L$ sets of parameters with a total number of $L(M+2)$. During the training of the MDN on the data set $D$, the parameters of the network are tuned in order to find the parameters which best fit the training data. 

The maximum likelihood principle is usually adopted. Equation~\ref{lm} indicates that maximizing the likelihood $L(\mathbf{m})$ equals to minimizing the misfit function $E(\mathbf{m})$. In the expression of the misfit function (Equation~\ref{em}), $P(\mathbf{m})$ is prior and independent of the observations and $k$ is a constant, making the last term $\log{[kP(\mathbf{m})]}$ a constant factor in the training process. Therefore, we can define a loss (or cost) function $Loss$ which quantifies how well the network fits the training data as
\begin{equation}
      \begin{aligned}
            Loss =& -{\rm log}P(\mathbf{m|d}) \\
            =& -\log\{  \sum_{l=1}^{L} \alpha_l(\mathbf{d}) \phi_l(\mathbf{m|d}) \}.
      \end{aligned}
      \label{eqloss}
\end{equation}
Once we find the network parameters which minimize the loss function $Loss$, we can compute the PPD which forms the solution of the inverse problem as a function of $\mathbf{m}$ using Equation~\ref{pmdmdn}.
\subsection{Derivation of geoacoustic statistics}

Suppose an appropriate designed MDN is trained now. We can obtain a multidimensional PPD with a formalism as Equation~\ref{pmdmdn} as soon as new observations data $\mathbf{d}$ are fed to the network. Known the multidimensional PPD, it is possible to compute the statistics in Equation~\ref{hatm}-\ref{cm} using numerical approaches. However, we propose to derive these statistics further using the MDN theory and make this process more efficient and straightforward.
\subsubsection{The MAP model}
The MAP model, defined in Equation~\ref{hatm}, involves locating the global maximum in the multidimensional parameter space. It is a nonlinear optimization problem and could be time-consuming when the dimensionality of $\mathbf{m}$ is large. From Equation~\ref{hatm}, \ref{pmdmdn} and \ref{phimd} we have
\begin{equation}
      \begin{aligned}
            \hat{\mathbf{m}} &= \operatorname*{argmax}\limits_{\mathbf{m}} P(\mathbf{m|d}) \\
            &= \frac{1}{(2\pi)^{M/2}} \operatorname*{argmax}\limits_{\mathbf{m}} 
            \{
                  \sum_{l=1}^{L} \frac{\alpha_l(\mathbf{d})}{\sigma_l(\mathbf{d})^M} 
                  {\rm exp} [-\frac{\left\| \mathbf{m}-\bm{\mu}_l(\mathbf{d})  \right\|^2}{2\sigma_l(\mathbf{d})^2}]
            \}.
      \end{aligned}
      \label{argmhat}
\end{equation}
Note that the exponential term in Equation~\ref{argmhat} is nonnegative and the maximum value of it is obtained when $\mathbf{m}=\bm{\mu}_l(\mathbf{d})$. If $L=1$, the solution is straightforward as $\mathbf{m}=\bm{\mu}_1(\mathbf{d})$. If $L>1$, it becomes more complicated. \citet{bishop1994mixture} provides a fast approximation under the assumption that the $L$ kernel functions are well separated from each other. The approximation to the MAP model is the center of the $l$th kernel $\bm{\mu}_l$ which satisfy
\begin{equation}
      \max\limits_l \{ \frac{\alpha_l(\mathbf{d})}{\sigma_l(\mathbf{d})^M} \}.
\end{equation}
It makes sense because when the kernel functions are well separated, the PPD is dominated by the kernel with the largest weight ${\alpha_l(\mathbf{d})}/{\sigma_l(\mathbf{d})^M}$, considering the maximum value of the exponential term in Equation~\ref{argmhat} is $1$ for all kernels.

\subsubsection{The mean model}
Substituting Equation~\ref{pmdmdn} and \ref{phimd} into Equation~\ref{overm} leads to
\begin{equation}
      \begin{aligned}
            \overline{\mathbf{m}} &= \int \mathbf{m} \sum_{l=1}^{L} \alpha_l(\mathbf{d}) \phi_l(\mathbf{m|d}) {\rm d}\mathbf{m} \\
            &= \sum_{l=1}^{L} \alpha_l(\mathbf{d}) \int \mathbf{m} \phi_l(\mathbf{m|d}) {\rm d}\mathbf{m}.
      \end{aligned}      
      \label{overmmdn}
\end{equation}
Note that $\phi_l(\mathbf{m|d})$ is a multivariate normal distribution, whose mean value can be written as
\begin{equation}
      \bm{\mu}_l(\mathbf{d}) = \int \mathbf{m} \phi_l(\mathbf{m|d}) {\rm d}\mathbf{m}.
      \label{mumdn}
\end{equation}
Using the relationship in Equation~\ref{mumdn}, Equation~\ref{overmmdn} can be simplified as
\begin{equation}     
      \overline{\mathbf{m}} =\sum_{l=1}^{L} \alpha_l(\mathbf{d}) \bm{\mu}_l(\mathbf{d}).           
      \label{meanmmdn}
\end{equation}
It indicates that the mean model derived from the multidimensional PPD is the weighted average of the centers of all the Gaussian kernels.
\subsubsection{The marginal PDFs}
1D and 2D marginal PDFs of model parameters are widely used by most researchers. If the output of the MDN is a multidimensional PPD, 1D, 2D and even higher dimensional marginal PDFs can be derived from it. Substituting Equations~\ref{pmdmdn} and \ref{phimd} into Equation~\ref{pmi} leads to
\begin{equation}
      \begin{aligned}
            P(m_i|\mathbf{d}) &= \int \delta(m_i-m_i^\prime)  P(\mathbf{m^\prime|d}) {\rm d}\mathbf{m^\prime} \\
            &= \frac{1}{(2\pi)^{M/2}} \int \delta(m_i-m_i^\prime)  
            \sum_{l=1}^{L} \frac{\alpha_l}{\sigma_l^M} 
                        {\rm exp} \{-\frac{\left\| \mathbf{m}^\prime-\bm{\mu}_l  \right\|^2}
                        {2\sigma_l^2}\}
            {\rm d}\mathbf{m^\prime} \\
            &= \frac{1}{(2\pi)^{M/2}} \sum_{l=1}^{L} \frac{\alpha_l}{\sigma_l^M} 
            \int \delta(m_i-m_i^\prime)              
                        {\rm exp} \{-\frac{\left\| \mathbf{m}^\prime-\bm{\mu}_l  \right\|^2}
                        {2\sigma_l^2}\} 
            {\rm d}\mathbf{m^\prime} \\
            &= \frac{1}{(2\pi)^{M/2}} \sum_{l=1}^{L} \frac{\alpha_l}{\sigma_l^M}I_l(m_i).
      \end{aligned}
      \label{pmimdn}
\end{equation}
Note that the $(\mathbf{d})$ is dropped from $\alpha_l(\mathbf{d})$, $\sigma_l(\mathbf{d})$ and $\bm{\mu_l(\mathbf{d})}$ for simplification. The integral term $I_l(m_i)$ can be computed as
\begin{equation}
      \begin{aligned}
            I_l(m_i) =& \int \delta(m_i-m_i^\prime)              
                        {\rm exp} \{-\frac{\left\| \mathbf{m}^\prime-\bm{\mu}_l  \right\|^2}
                        {2\sigma_l^2}\}
                  {\rm d}\mathbf{m^\prime} \\
                  =& \int \delta(m_i-m_i^\prime)
                        {\rm exp} \{ -\frac{1}{2\sigma_l^2}[
                              (m_1^\prime-\mu_{l1})^2 + (m_2^\prime-\mu_{l2})^2 + \cdots  \\
                        & + (m_i^\prime-\mu_{li})^2 + \cdots + (m_M^\prime-\mu_{lM})^2                              
                        ]
                        \}
                        {\rm d}m_1^\prime {\rm d}m_2^\prime \cdots {\rm d}m_i^\prime \cdots {\rm d}m_M^\prime \\
                  =& \int {\rm exp} \{ -\frac{1}{2\sigma_l^2} (m_1^\prime-\mu_{l1})^2 \} {\rm d}m_1^\prime  \int {\rm exp} \{ -\frac{1}{2\sigma_l^2} (m_2^\prime-\mu_{l2})^2 \} {\rm d}m_2^\prime \cdots  \\
                  & \int \delta(m_i-m_i^\prime) {\rm exp} \{ -\frac{1}{2\sigma_l^2} (m_i^\prime-\mu_{li})^2 \} {\rm d}m_i^\prime \cdots \int {\rm exp} \{ -\frac{1}{2\sigma_l^2} (m_M^\prime-\mu_{lM})^2 \} {\rm d}m_M^\prime \\
                  =& \sqrt{2\pi}\sigma_l \cdot \sqrt{2\pi}\sigma_l \cdots {\rm exp} \{ -\frac{1}{2\sigma_l^2} (m_i-\mu_{li})^2 \} \cdots \sqrt{2\pi}\sigma_l \\
                  =& (2\pi)^{(M-1)/2}\sigma_l^{M-1} {\rm exp} \{ -\frac{1}{2\sigma_l^2} (m_i-\mu_{li})^2 \},
      \end{aligned}
      \label{ilmi}
\end{equation}
in which the relationship $\int {\rm exp} \{ -\frac{1}{2\sigma_l^2} (m^\prime-\mu_{li})^2 \} {\rm d}m^\prime = \sqrt{2\pi}\sigma_l$ is adopted. This relationship is proved in APPENDIX~\ref{apif}. Substituting Equation~\ref{ilmi} back into Equation~\ref{pmimdn} we have
\begin{equation}
      P(m_i|\mathbf{d}) = \frac{1}{\sqrt{2\pi}} \sum_{l=1}^{L} \frac{\alpha_l}{\sigma_l} {\rm exp} \{ -\frac{1}{2\sigma_l^2} (m_i-\mu_{li})^2 \}.
      \label{pmidfinal}
\end{equation}

The derivation for 2D joint marginal PDF is similar to 1D marginal PDF and is omitted. Similar to 1D marginal PDF (Equation~\ref{pmidfinal}), 2D joint marginal PDF can be derived as

\begin{equation}
      \begin{aligned}
            P(m_i,m_j|\mathbf{d}) &= \int \delta(m_i-m_i^\prime)\delta(m_j-m_j^\prime)  P(\mathbf{m^\prime|d}) {\rm d}\mathbf{m^\prime} \\
            &= \frac{1}{2\pi} \sum_{l=1}^{L} \frac{\alpha_l}{\sigma_l^2} \exp \{ -\frac{1}{2\sigma_l^2} [(m_i-\mu_{li})^2+(m_j-\mu_{lj})^2] \}.
      \end{aligned}
      \label{pmimjmdnfinal}
\end{equation}

Based on Equations~\ref{pmidfinal} and \ref{pmimjmdnfinal}, the formulas to compute higher dimensional joint marginal PDFs are straightforward. When the dimensionality reaches $M$, Equation~\ref{pmimjmdnfinal} converges to Equation~\ref{pmdmdn}.

\subsubsection{The model covariance matrix}
The diagonal elements of $\mathbf{C}_\mathbf{m}$ (Equation~\ref{cm}) is computed as

\begin{equation}
      \begin{aligned}
            \mathbf{C}_{m_{ii}} =& \int (m_i-\overline{m}_i)^2 P(\mathbf{m|d}) {\rm d}\mathbf{m} \\
            =& \frac{1}{(2\pi)^{M/2}} \sum_{l=1}^{L} \frac{\alpha_l}{\sigma_l^M}
            \int (m_i-\overline{m}_i)^2              
                        {\rm exp} \{-\frac{\left\| \mathbf{m}-\bm{\mu}_l  \right\|^2}
                        {2\sigma_l^2}\} 
            {\rm d}\mathbf{m} \\
            =& \frac{1}{(2\pi)^{M/2}} \sum_{l=1}^{L} \frac{\alpha_l}{\sigma_l^M} 
            \int {\rm exp} \{ -\frac{1}{2\sigma_l^2} (m_1-\mu_{l1})^2 \} {\rm d}m_1   \int {\rm exp} \{ -\frac{1}{2\sigma_l^2} (m_2 -\mu_{l2})^2 \} {\rm d}m_2  \cdots  \\
            & \int {(m_i-\overline{m}_i)^2}  {\rm exp} \{ -\frac{1}{2\sigma_l^2} (m_i -\mu_{li})^2 \} {\rm d}m_i  \cdots \int {\rm exp} \{ -\frac{1}{2\sigma_l^2} (m_M -\mu_{lM})^2 \} {\rm d}m_M  \\
            =& \frac{1}{(2\pi)^{M/2}} \sum_{l=1}^{L} \frac{\alpha_l}{\sigma_l^M} 
            \sqrt{2\pi}\sigma_l \cdot \sqrt{2\pi}\sigma_l \cdots \sqrt{2\pi}\sigma_l[(\overline{m}_i-\mu_{li})^2+\sigma_l^2] \cdots \sqrt{2\pi}\sigma_l \\
            =& \sum_{l=1}^{L} \alpha_l [(\overline{m}_i-\mu_{li})^2+\sigma_l^2],
      \end{aligned}
      \label{cmmdn}
\end{equation}
in which the relationship $\int {(m_i-\overline{m}_i)^2}  {\rm exp} \{ -\frac{1}{2\sigma_l^2} (m_i -\mu_{li})^2 \} {\rm d}m_i  = \sqrt{2\pi}\sigma_l[(\overline{m}_i-\mu_{li})^2+\sigma_l^2]$ is adopted. This relationship is proved in APPENDIX~\ref{apif1}. 

The off-diagonal elements of $\mathbf{C}_\mathbf{m}$ (Equation~\ref{cm}) is computed as

\begin{align}
      \mathbf{C}_{m_{ij}} =& \int (m_i-\overline{m}_i)(m_j-\overline{m}_j) P(\mathbf{m|d}) {\rm d}\mathbf{m} \notag \\
      =& \frac{1}{(2\pi)^{M/2}} \sum_{l=1}^{L} \frac{\alpha_l}{\sigma_l^M}
      \int (m_i-\overline{m}_i)(m_j-\overline{m}_j)              
                  {\rm exp} \{-\frac{\left\| \mathbf{m}-\bm{\mu}_l  \right\|^2}
                  {2\sigma_l^2}\} 
      {\rm d}\mathbf{m} \notag \\
      =& \frac{1}{(2\pi)^{M/2}} \sum_{l=1}^{L} \frac{\alpha_l}{\sigma_l^M} 
      \int {\rm exp} \{ -\frac{1}{2\sigma_l^2} (m_1-\mu_{l1})^2 \} {\rm d}m_1   \int {\rm exp} \{ -\frac{1}{2\sigma_l^2} (m_2 -\mu_{l2})^2 \} {\rm d}m_2  \cdots \notag \\
      & \int {(m_i-\overline{m}_i)}  {\rm exp} \{ -\frac{1}{2\sigma_l^2} (m_i -\mu_{li})^2 \} {\rm d}m_i  \cdots \\
      & \int {(m_j-\overline{m}_j)}  {\rm exp} \{ -\frac{1}{2\sigma_l^2} (m_j -\mu_{lj})^2 \} {\rm d}m_j \cdots  \int {\rm exp} \{ -\frac{1}{2\sigma_l^2} (m_M -\mu_{lM})^2 \} {\rm d}m_M \notag \\
      =& \frac{1}{(2\pi)^{M/2}} \sum_{l=1}^{L} \frac{\alpha_l}{\sigma_l^M}
      \sqrt{2\pi}\sigma_l \cdot \sqrt{2\pi}\sigma_l \cdots \sqrt{2\pi}\sigma_l(\mu_{li}-\overline{m}_i) \cdots \sqrt{2\pi}\sigma_l(\mu_{lj}-\overline{m}_j) \cdots \sqrt{2\pi}\sigma_l \notag \\
      =& \sum_{l=1}^{L} \alpha_l (\mu_{li}-\overline{m}_i) (\mu_{lj}-\overline{m}_j). \notag
\label{cmmijdn}
\end{align}


Then the inter-parameter correlation matrix can be computed using Equation~\ref{correlationmatrix}.

\section{The MDN inversion}
In this section we perform the MDN inversion of shear-wave velocity profile of seabed from surface wave dispersion curves. Surface wave measurements are usually recorded directly in an active source survey \citep{Dong2011} or reconstructed from passive collected ambient noise \citep{Wu2019}. Surface wave dispersion curves can be extracted from these measurements and used to estimate the shear-wave velocity profile \citep{Li2012}. The method described here is general and can be easily applied to other geoacoustic inverse problems.

\subsection{Preparation of the training set}

The training of machine learning algorithms relies on sufficient training data, especially for problems with high complexity. However, only limited observational data or even no geophysical ground-truth data are available in many geoacoustic problems. Hence, training the network on the simulation data is a good alternative. The forward problem is solved using DISPER80 subroutines developed by \citet{Saito1980DISPER80}.
\subsubsection{Model parametrization}
\begin{figure} 
      \centering
      \includegraphics[width=0.7\textwidth]{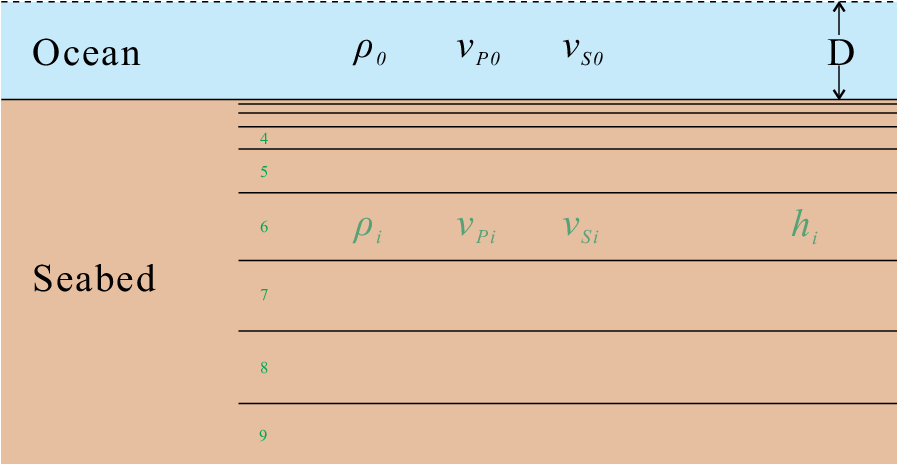}
      \caption{Model parametrization. The bottom layer is a semi-infinite layer. Note that layers with different thickness do not keep the same scale for display purposes.}
      \label{figlayermodel}
\end{figure}
\begin{figure} 
      \centering
      \includegraphics[width=0.5\textwidth]{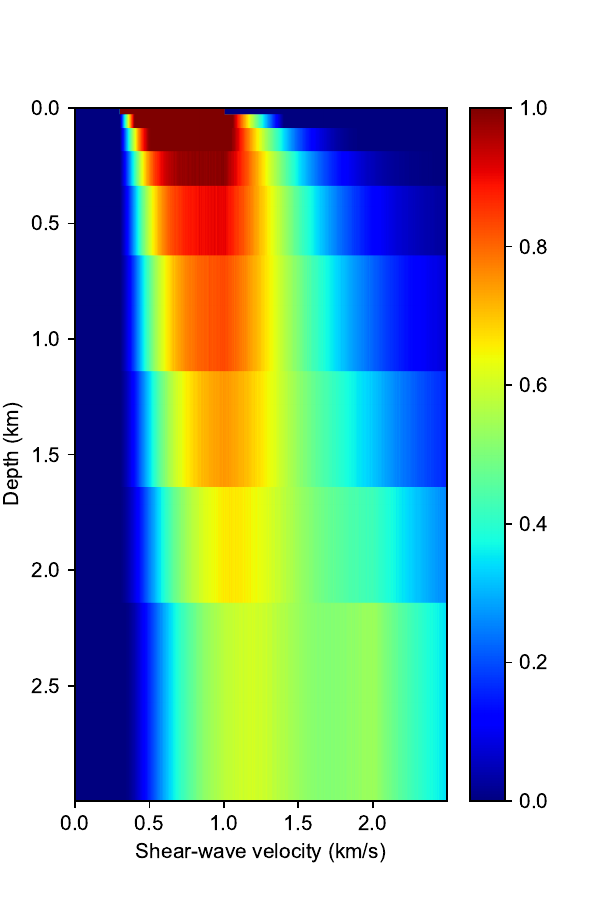}
      \caption{The shear-wave velocity density distribution.}
      \label{figvsdd}
\end{figure} 
The shear-wave velocity profile is parametrized as a layered structure, which is shown in Figure~\ref{figlayermodel}.  The symbols $\rho$, $v_P$, $v_S$ and $h$ represent the density, compressional wave velocity, shear wave velocity and denpth of a layer. The water layer is colored in blue, and the seabed layers are colored in darksalmon. $\rho_0=1.0$ kg/m$^3$, $v_{P0}=1.49$ km/s, $v_{S0}=0$ and $D=0.325$ km are used for water layer. The seabed is composed of 9 layers with $\mathbf{h}= (0.325, 0.03, 0.06, 0.10, 0.15, 0.30, 0.50, 0.50, 0.50, inf)$ m. The first top layer is a near-surface layer which is thought to be soft and unconsolidated with $v_{S1} \sim U(0.3~\rm{km/s}, 1.0~\rm{km/s})$. \citet{Galetti2017} found that DISPER80 produces unreliable result if a large velocity drop exists between two consecutive layers with increasing depth in the model. Therefore, we define the constraint on the velocity model as
\begin{equation}
      \begin{aligned}
            & v_{Si} \sim U(lbound,ubound),~i=2,3,\cdots,9 \\
            & lbound = \max(0.8v_{Si-1},v_{S1}) \\
            & ubound = \min(1.4v_{Si-1},2.5~\rm{km/s}).            
      \end{aligned}
      \label{vsiu}
\end{equation}
The constraint makes the shear-wave velocity of each layer at least 80 and at most 140 percent of the upper consecutive layer. Implicitly a hard bound ($v_{s1}$, 2.5 km/s) is implied in the constraint. The constraint excludes lots of unusual models in the real world. Since the dispersion property of the surface wave is not sensitivity to the density $\rho$ and the compressional velocity $v_P$, $\rho$ and $v_P$ are computed using the empirical relations \citep{Castagna1985, Brocher2005}
\begin{equation}
      v_P = 1.16 v_S + 1.36,
\end{equation}
and
\begin{equation}
      \rho = 1.74v_P^{0.25}.
\end{equation}

$10^7$ model-samples are generated randomly and the shear-wave velocity density distribution is shown in Figure~\ref{figvsdd}. These samples are well distributed based on our definition and most possible models in the real world are covered. Based on the definition of Equation~\ref{vsiu}, the velocity of a model more likely tend to increase with depth because of the setting of the bounds. $10^7$ corresponding phase-velocity dispersion curves (observations) are computed using the forward model DISPER80. This procedure is fully paralleled using parallel computing technique. Thus we obtain a data set $D = \{(\mathbf{d}_i,\mathbf{m}_i):i = 1,...,N_D\}$ with $N_D = 10^7$. 

\subsubsection{Uncertainties of the observations}

The realistic measurements of phase-velocity dispersion curves are subject to uncertainties. We assume that the uncertainties of the observations are independent, Gaussian-distributed random processes and add synthetic noise to the simulated observations according to the probability density
\begin{equation}
      \rho(\widetilde{\mathbf{d}})=\frac{1}{(2 \pi)^{M / 2}\left|\mathbf{C}_{D}\right|^{1 / 2}} \exp \left\{-\frac{1}{2}\left( \mathbf{d}-\widetilde{\mathbf{d}}\right)^{T} \mathbf{C}_{D}^{-1}\left( \mathbf{d}-\widetilde{\mathbf{d}}\right)\right\},
      \label{eqrhod}
\end{equation}
where $\widetilde{\mathbf{d}}$ denotes the noisy observations and $\mathbf{C}_{D}$ denotes the diagonal data covariance matrix. The diagonal elements $\sigma_i^2$ of $\mathbf{C}_{D}$ are defined as
\begin{equation}
      \sigma_i = \epsilon d_i,
\end{equation} 
where $\epsilon$ is the noise intensity control factor and 0.05 is used in this paper.

\subsection{Regularization}

The capacity of the MDN to approximate the nonlinear mapping of inverse problems relies on sufficient flexibility of the network. In practice, we are prone to adding plenty of freedoms to the network to make sure that it is sufficiently flexible to model the nonlinear mapping with little bias. However, too great flexibility can result in the overfitting problem, which means that the network fits the training set perfectly but show bad performance on new data. Regularization is needed to constrain the network.

Adding noise to the observations works like an implicit regularizer. It prevents the network from learning small variations in the observations and makes the MDN less sensitive to variations within uncertainties. \citet{bishop1994mixture} and \citet{Meier2007a} show that adding noise to the observations has the same effect as using a regularized error function which involves a penalty term. 

Besides, an early stopping approach is employed in this paper to prevent overfitting. The model is evaluated on the validation set at regular checkpoints and the value of the loss function is compared with the previously lowest one. The value of the \q{winner} is updated after each comparison. If the value stops updating for a certain time, the training of the network is then interrupted and the latest \q{winner} model is adopted.



\subsection{Network design}
\begin{figure} 
      \centering
      \includegraphics[width=\textwidth]{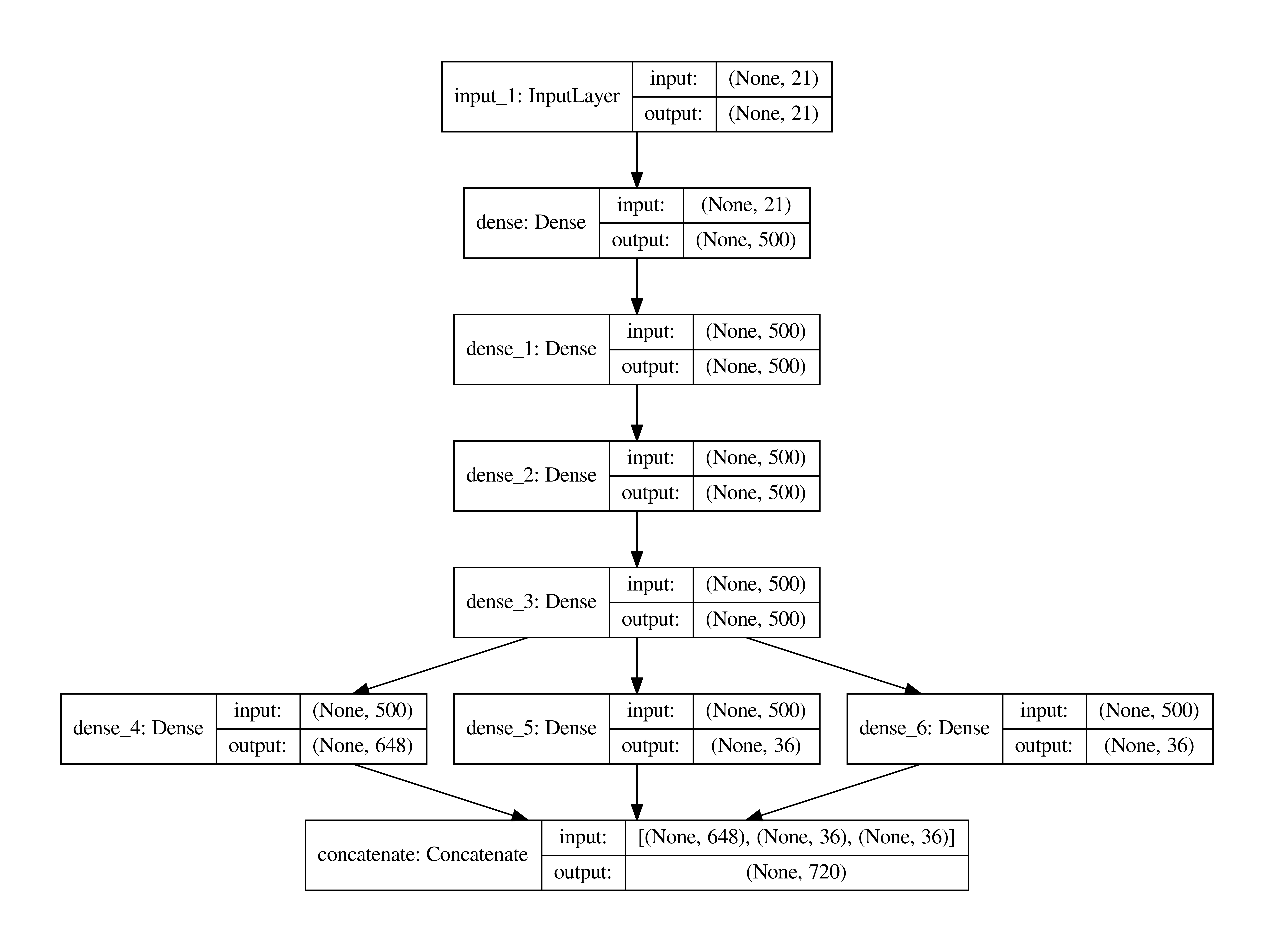}
      \caption{Network structure. Dense layer is the regular deeply connected neural network layer.}
      \label{fignetwork}
\end{figure}

A 7-layer MDN is designed for the inversion of shear-wave velocity problem, including an input layer, an output layer and 5 dense layers (the 5th dense layer is composed of 3 branches). The basic shape of the network is shown in Figure~\ref{fignetwork}. The input layer takes the discretized points of the phase-velocity dispersion curves as inputs. The dense\_4, dense\_5 and dense\_6 layers respectively compute $\bm{\mu}$, $\sigma$ and $\alpha$ of all kernels which fully define the multidimensional PPD. The last layer concatenates all the outputs and performs as an output layer. Once trained, the statistics of shear-wave velocities for all layers of the seabed can be extracted from a single MDN. Table~\ref{tabparacount} displays the distribution of trainable parameters in the network. The total number of trainable parameters included in the network is 960,896.

\begin{figure} 
      \centering
      \includegraphics[width=0.6\textwidth]{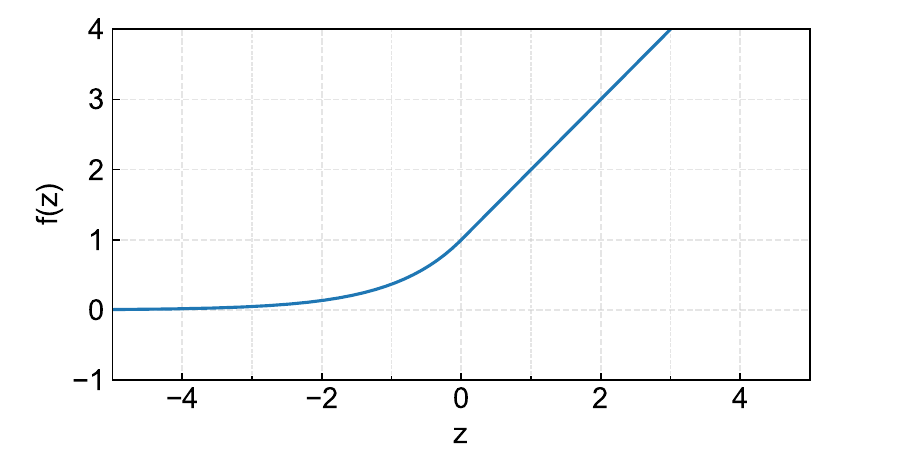}
      \caption{The modified ELU activation function.}
      \label{figmodifiedelu}
\end{figure}

The ReLU activation function is used in the first 4 dense layers. $\bm{\mu}$ is a location parameter which denotes the center of a kernel. The output of dense\_4 layer is directly used as $\bm{\mu}$ and no activation function is used. $\alpha$ is a mixing coefficient which should satisfy the constraint of Equation~\ref{sumsigma1}. Thus the softmax activation function is used in dense\_5 layer for the calculation of $\alpha$. $\sigma$ is a scale parameter which denotes the variance of a kernel. A modified ELU activation function $f(z) = ELU(1,z)+1$ \citep{brando2017mixture} is used in dense\_5 layer. $f(z)$ is smooth and positive everywhere (shown in Figure~\ref{figmodifiedelu}) and is suitable for the activation of $\sigma$.

\begin{table}
      \centering
      \caption{Network summary.}
      \label{tabparacount}
      \begin{tabular}{lllll}
      \toprule
      Layer (type)              & Param \# & Output shape & Activation function &   \\
      \midrule
      input\_1 (InputLayer)     & 0        & (None, 21)   & None                &  \\
      dense (Dense)             & 11000    & (None, 500)  & ReLU                &  \\
      dense\_1 (Dense)          & 250500   & (None, 500)  & ReLU                &  \\
      dense\_2 (Dense)          & 250500   & (None, 500)  & ReLU                &  \\
      dense\_3 (Dense)          & 250500   & (None, 500)  & ReLU                &  \\
      dense\_4 (Dense)          & 324648   & (None, 324)  & None                &  \\
      dense\_5 (Dense)          & 18036    & (None, 36)   & Softmax             &  \\
      dense\_6 (Dense)          & 18036    & (None, 36)   & Modified ELU        &  \\
      concatenate (OutputLayer) & 0        & (None, 396)  & None                &  \\
      Total params:             & 960,896    &         & &  \\
      Trainable params:         & 960,896    &         & &  \\
      Non-trainable params:     & 0          &         & &  \\
      \bottomrule
      \end{tabular}
\end{table}

\subsection{Network training}
The network is trained on the data set $D = \{(\mathbf{d}_i,\mathbf{m}_i):i = 1,...,N_D\}$ with $10^7$ samples. The data set is shuffled and split into a training set, a validation set and a test set with 9,800,000, 100,000 and 100,000 samples, respectively. 

The network is trained on a server with a machine configuration as
\begin{enumerate}
      \item[-] Operating system: Ubuntu 18.04.4 LTS
      \item[-] CPU: Intel Xeon(R) E5-1620 v4 @ 3.50GHz $\times$ 8
      \item[-] GPU: GeForce GTX 1080 Ti
      \item[-] Memory: 64 GB
      \item[-] GPU memory: 12 GB  
      \item[-] Hard disk: Samsung SSD 850 PRO 512 GB 
\end{enumerate}
The following programming tools and software libraries were used to implement the experiment.
\begin{enumerate}
      \item[-] Python 3.7.7
      \item[-] Tensorflow-GPU 1.13.1
      \item[-] Keras 2.3.1 
\end{enumerate}
In this study it takes about 8.75 hours to run 1500 epochs of the training with batch size equals to 8192.

\section[]{Results}
\subsection{Network evaluation}
The designed MDN is trained on the prepared training set and evaluated on the validation set during the training process. The noise is added to the two sets based on Equation~\ref{eqrhod} to form a noisy training. The maximum value of training epoch is set as 1500.

\subsubsection{The loss function}

\begin{figure} 
      \centering
      \includegraphics[width=\textwidth,trim=30 0 30 0,clip]{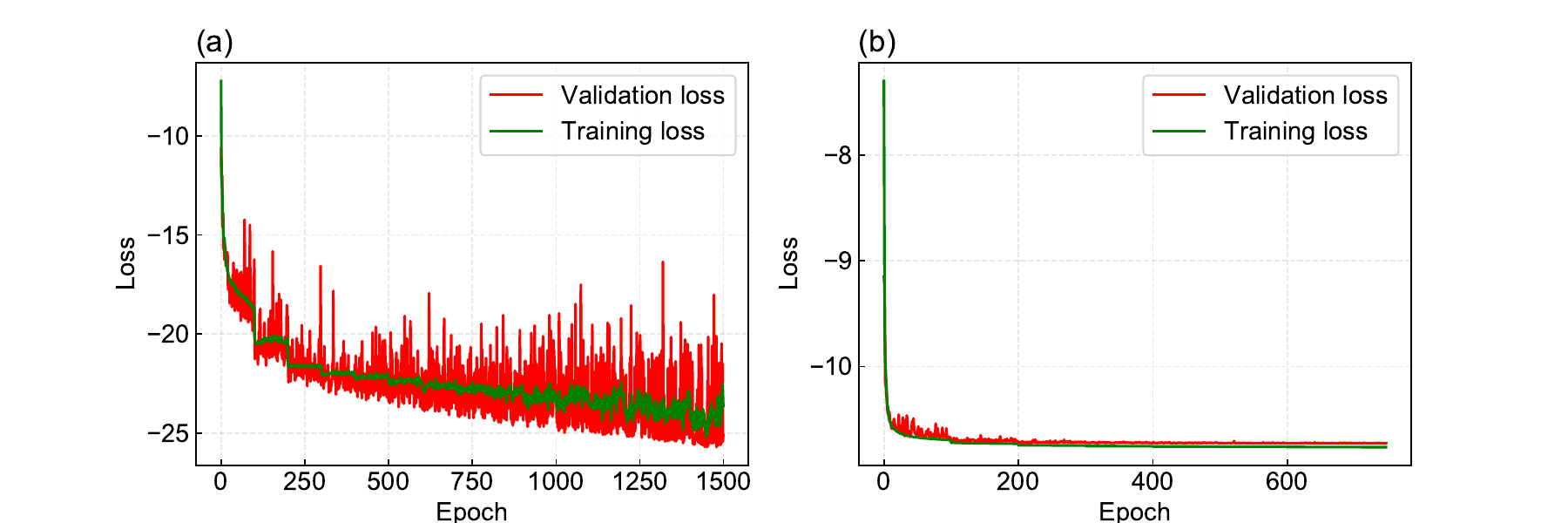}
      \caption{The loss as a function of epoch. (a) The loss of the training on noiseless data. (b) The loss of the training on noisy data.}
      \label{figloss}
\end{figure}

Figure~\ref{figloss} compares the loss (Equation~\ref{eqloss}) of noiseless and noisy training. Figure~\ref{figloss}a shows that the validation loss of the noiseless training can reach a minimum of around -25.72 at epoch=1480. However, the loss curve is oscillating at a large scale during the training process, which indicates that the network is sensitive to slight changes and may be overfitting the training set. Figure~\ref{figloss}b shows a better convergence of the noisy training compared with Figure~\ref{figloss}a. The validation loss reaches a minimum of around -10.73 at epoch=648. A small oscillation occurs at the first 100 epoches but it comes back to convergence soon. At epoch=748, the early stopping condition is triggered and the training process is stopped. The performance of the network in Figure~\ref{figloss}b is similar on the training set and validation set, indicating that overfitting does not happen in the training.

\subsubsection{The predictions versus true velocity models}

\begin{figure} 
      \centering
      \includegraphics[width=0.8\textwidth,trim=100 230 100 360,clip]{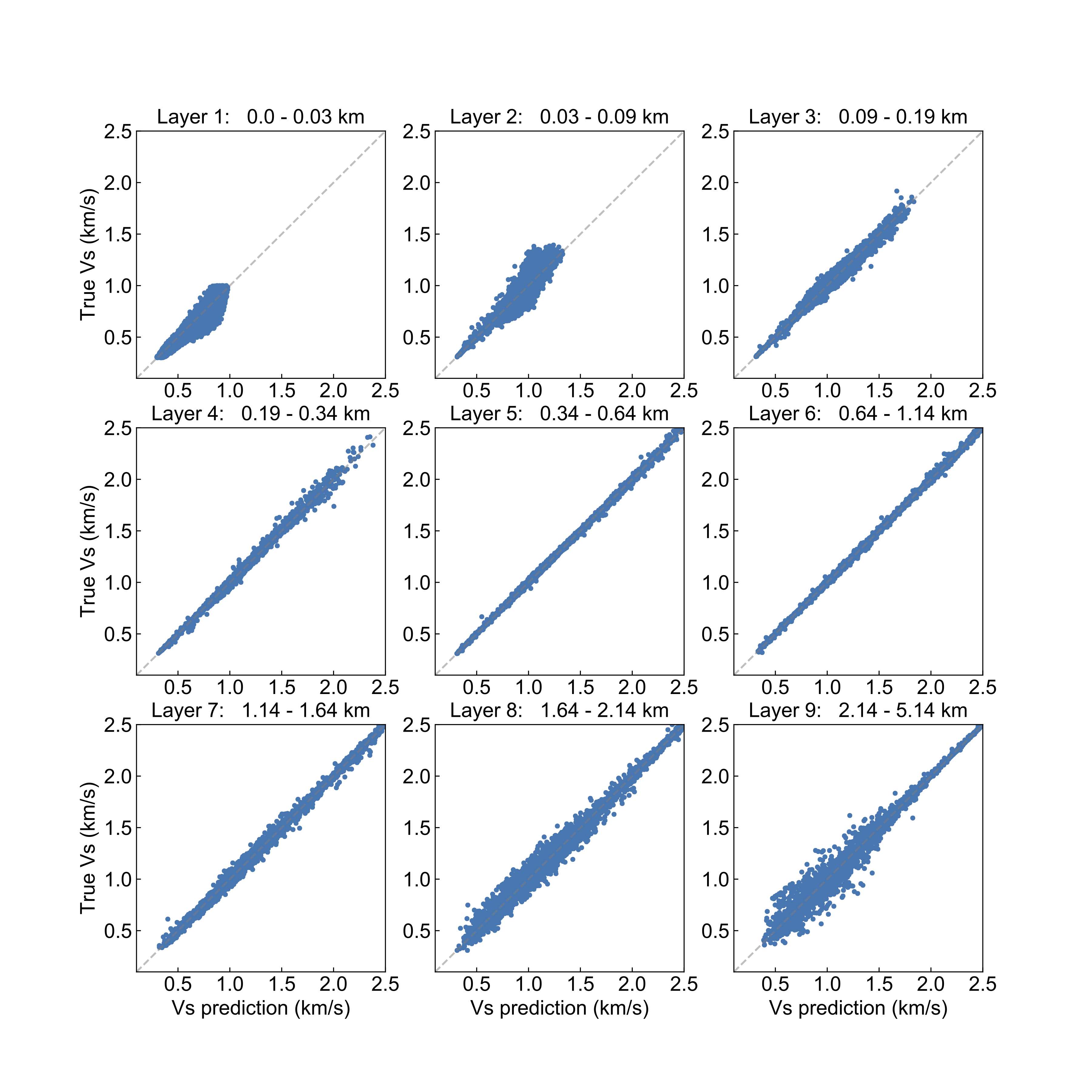} 
      \caption{The predictions of noiseless training on noiseless test set versus true velocity models.}
      \label{figpredictionVStrue_nlnl}
\end{figure}

\begin{figure} 
      \centering
      \includegraphics[width=0.8\textwidth,trim=100 230 100 360,clip]{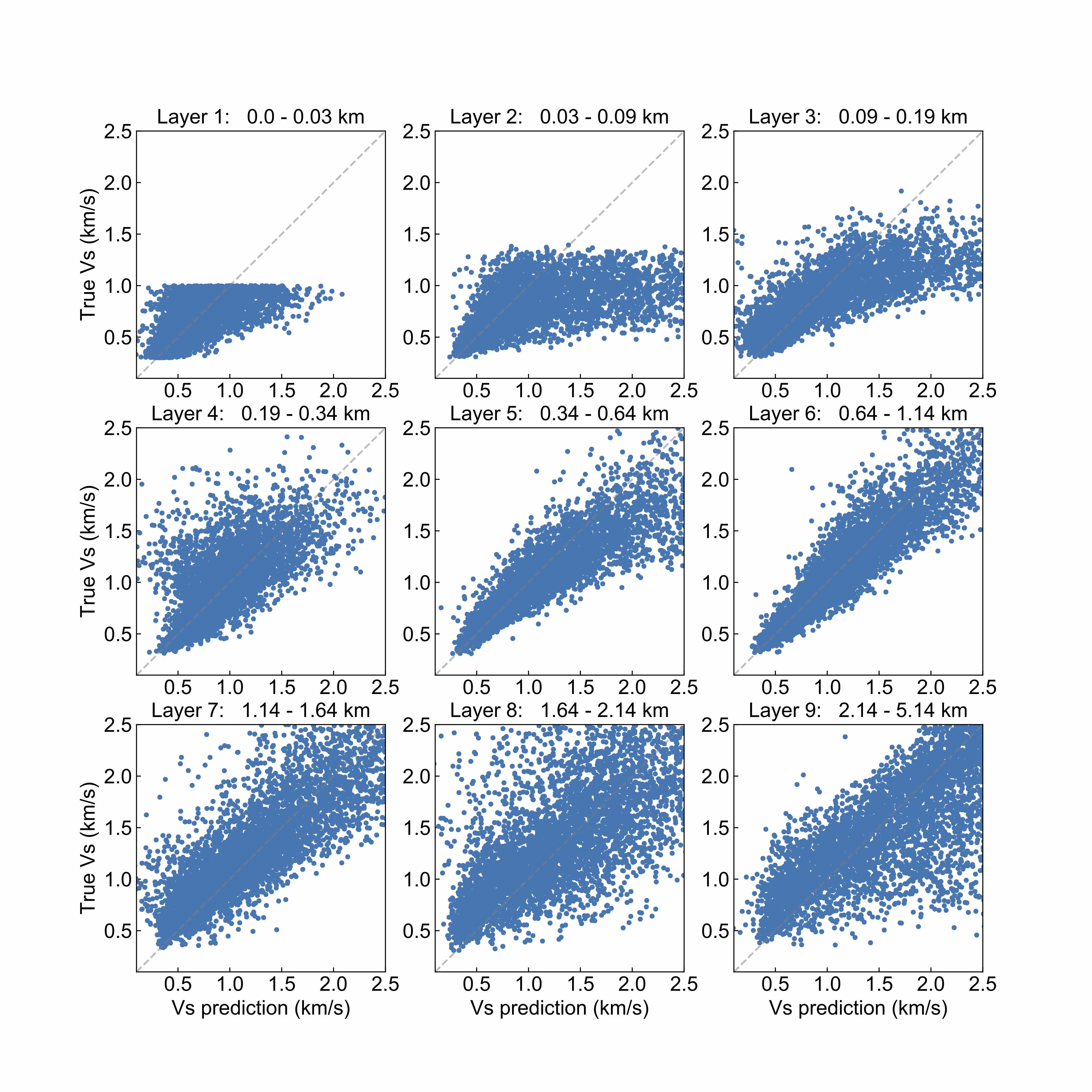} 
      \caption{The predictions of noiseless training on noisy test set versus true velocity models.}
      \label{figpredictionVStrue_nlny}
\end{figure}

\begin{figure} 
      \centering
      \includegraphics[width=0.8\textwidth,trim=100 230 100 360,clip]{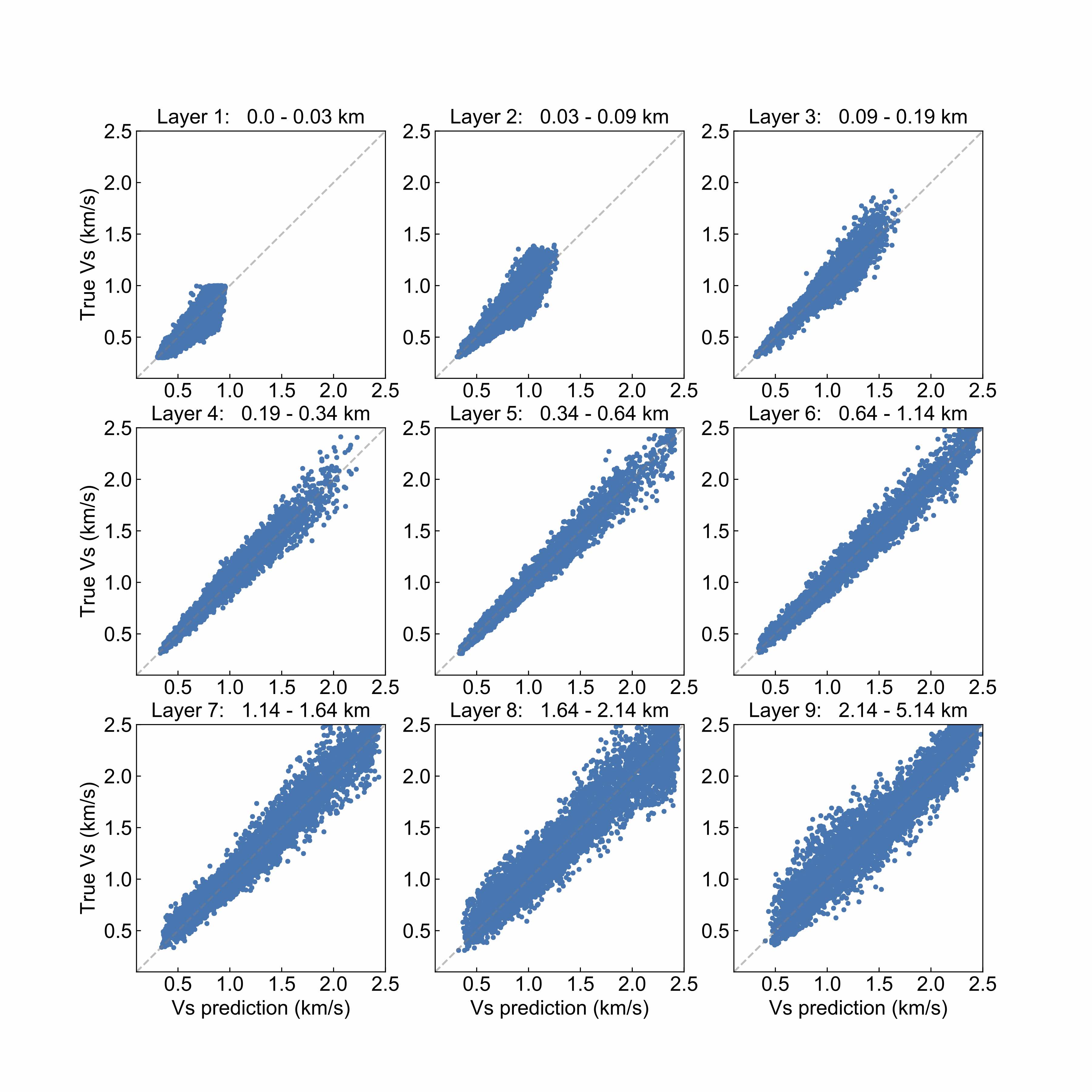} 
      \caption{The predictions of noisy training on noisy test set versus true velocity models.}
      \label{figpredictionVStrue_nyny}
\end{figure}

The trained networks are evaluated on the noiseless test set and noisy test set. Random noise is added to the noisy test set based on Equation~\ref{eqrhod} to simulate the measurement errors in the real world. Since the test set is never seen by the networks during the training process, the performance of the networks on the test set is a good indicator for generalization.

The predictions (mean velocity model) of different networks are plotted against the true velocity models in Figures~\ref{figpredictionVStrue_nlnl}, \ref{figpredictionVStrue_nlny} and \ref{figpredictionVStrue_nyny}. The predictions are computed using Equation~\ref{meanmmdn}. Theoretically, precise predictions should fall on the diagonal lines of each subplot.

Figure~\ref{figpredictionVStrue_nlnl} presents the predictions of noiseless training on noiseless test set. Generally speaking, the predictions of all layers perform very well with corresponding dots locating closely to the diagonal lines. Layers 1 and 2 are very thin and may contribute relatively less to the observation (dispersion curve), making the loss function being less sensitive to the velocities of layers 1 and 2. As a result, the predictions of layers 1 and 2 perform slightly worse. Comparing layers 6 to 9 we can find that the performance becomes worse when the depth goes deeper, indicating larger uncertainties for deeper layers. It is because the penetration ability of the surface wave attenuates as the depth becomes larger. It determines that the sensitivity of the dispersion data to the deeper layer is lower than that of the shallow layer, so the prediction performance of deeper layers is also worse than the shallower Layers.

Figure~\ref{figpredictionVStrue_nlny} presents the predictions of noiseless training on noisy test set. The overall performance of the network is very bad. The reason is that the network overfit noiseless training set, making it very sensitive to slight fluctuation or error of the observation and resulting in bad generalization performance. When the observation data is accurate, a highly accurate prediction can be obtained (Figure~\ref{figpredictionVStrue_nlnl}). But when the observation data contains uncertainties, its prediction effect deteriorates sharply.

Figure~\ref{figpredictionVStrue_nyny} presents the predictions of noisy training network on noisy test set. It shows improved performance at every layer compared with Figure~\ref{figpredictionVStrue_nlny}. Similar to Figure~\ref{figpredictionVStrue_nlnl}, the predictions of the first two layers perform slightly worse due to small thickness and the uncertainty of the prediction increases gradually as the depth goes deeper due to the penetration ability of the surface wave.

\begin{table}
      \centering
      \caption{Final loss values on validation set and test set.}
      \label{tablosscp}
      \begin{tabular}{lllll}
      \toprule
      \multicolumn{1}{c}{Category} &
      \multicolumn{1}{c}{\begin{tabular}[c]{@{}c@{}}location of \\ minimum (epoch) \end{tabular}} &
      \multicolumn{1}{c}{\begin{tabular}[c]{@{}c@{}} loss on \\ validation set \end{tabular}} &
      \multicolumn{1}{c}{\begin{tabular}[c]{@{}c@{}} loss on \\ noiseless test set \end{tabular}} &
      \multicolumn{1}{c}{\begin{tabular}[c]{@{}c@{}} loss on \\ noisy test set \end{tabular}} \\
      \midrule
      \multicolumn{1}{c}{Noiseless training network} & \multicolumn{1}{c}{1480} & \multicolumn{1}{c}{-25.72}  & \multicolumn{1}{c}{-25.69} & \multicolumn{1}{c}{14808.89} \\
  \multicolumn{1}{c}{Noisy training network} & \multicolumn{1}{c}{648}  & \multicolumn{1}{c}{-10.73} & \multicolumn{1}{c}{-12.73} & \multicolumn{1}{c}{-10.71} \\
      \bottomrule
      \end{tabular}
  \end{table}

Table~\ref{tablosscp} lists the comparison of the final loss values for different networks on different sets. On the noiseless test set, the performance of the noiseless training network is close to that on the validation set. However, on the noisy test set, the performance of the noiseless training network is poor, and its loss is greatly increased relative to the noiseless situation, indicating that the generalization performance of this network is poor. On the noiseless test set, the performance of the noisy training network is slightly improved compared to the validation set. On the noisy test set, the performance of the noisy training network is close to that on the validation set, which shows that this network has better generalization performance. In general, the network trained on noisy data is more robust and generalizes better. Therefore, the network trained on noisy data is chosen use in the following synthetic cases and real data cases.

Note that the points off the diagonal line are not necessarily bad predictions because they may represent other alternative solutions when non-unique solutions exist for the problem.

\subsection{Synthetic cases}

\begin{figure} 
      \centering
      \includegraphics[width=0.8\textwidth,trim=70 80 70 100,clip]{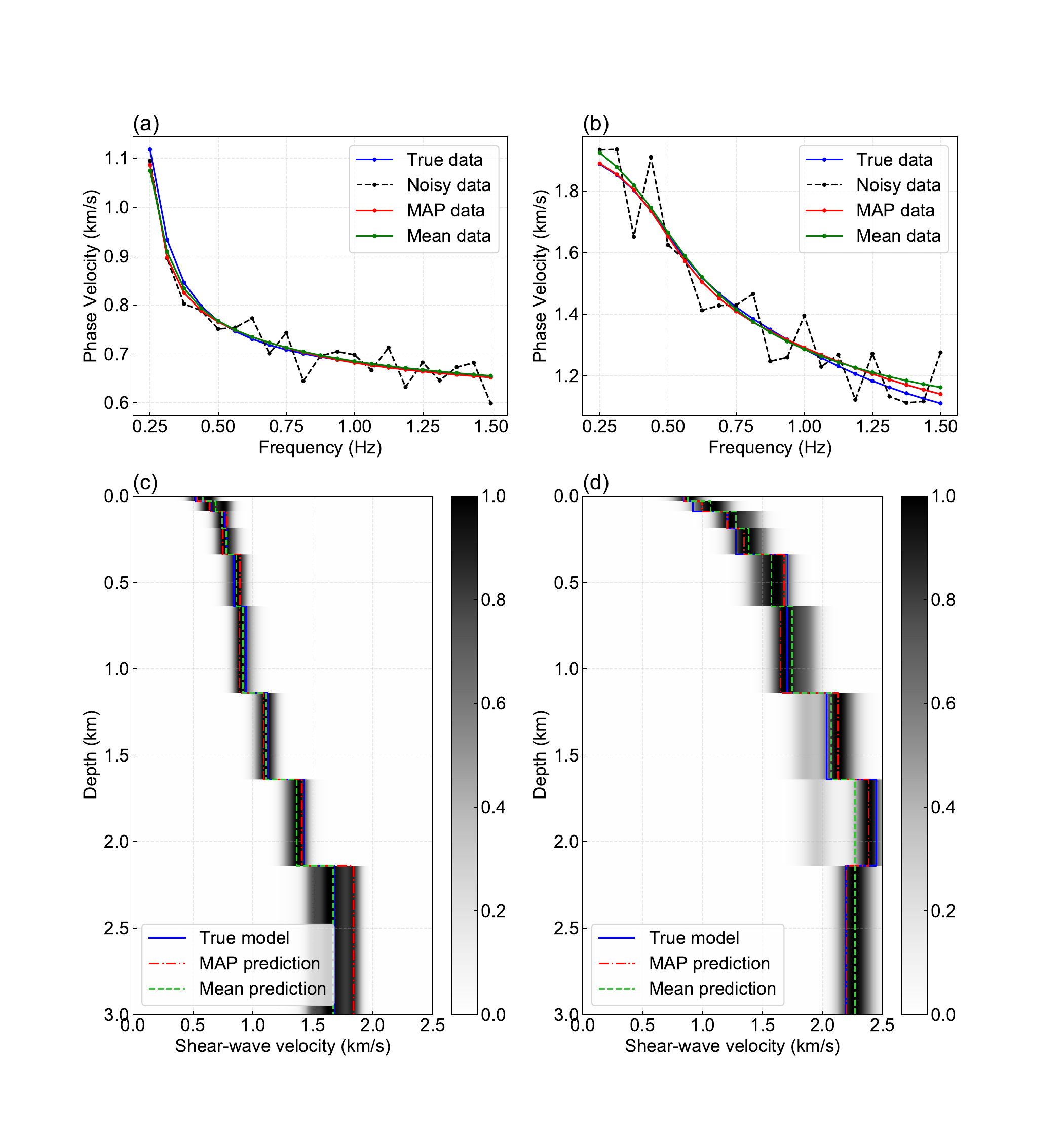} 
      \caption{The performance of noisy training network on synthetic cases.}
      \label{figpdfs_9_1}
\end{figure}

\begin{figure} 
      \centering
      \includegraphics[width=0.8\textwidth,trim=70 80 70 100,clip]{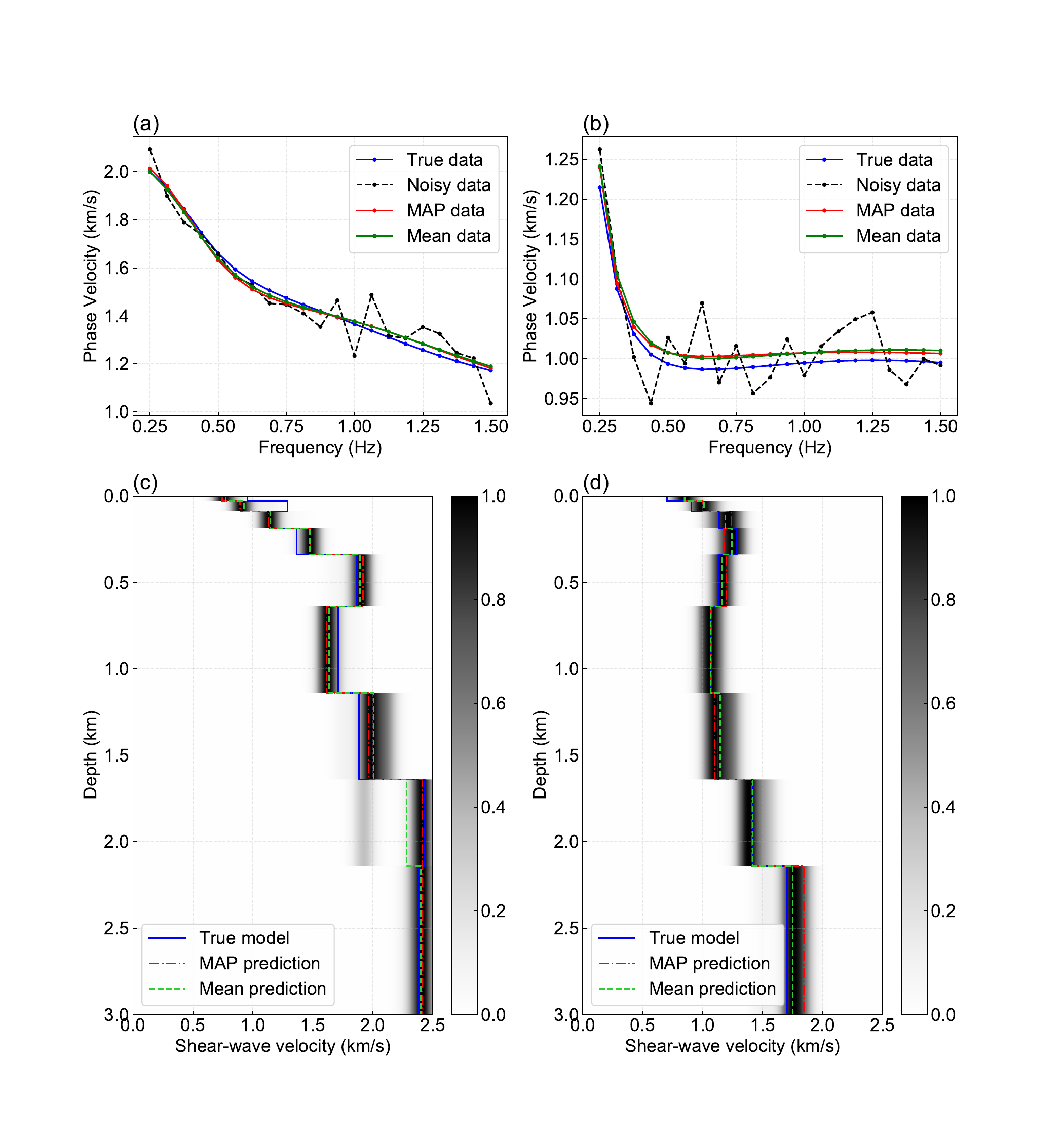} 
      \caption{The performance of noisy training network on synthetic cases.}
      \label{figpdfs_9_2}
\end{figure}

To illustrate how the predictions fit the true model and data, we take several synthetic cases for example. 

Figure~\ref{figpdfs_9_1} presents two synthetic cases. The true dispersion curves are displayed in Figure~\ref{figpdfs_9_1}a and \ref{figpdfs_9_1}b with blue solid lines. Noise is added to the dispersion curves using Equation~\ref{eqrhod} and the noisy dispersion curves are displayed in Figure~\ref{figpdfs_9_1}a and \ref{figpdfs_9_1}b with black dashed lines. The noisy dispersion curves in Figure~\ref{figpdfs_9_1}a and \ref{figpdfs_9_1}b are fed to the MDN and we can get the multidimensional PPD as outputs. Then the mean prediction models and the MAP prediction models are computed using Equations \ref{argmhat} and \ref{meanmmdn}, and the 1D PPDs are computed using Equation \ref{pmidfinal}. These predictions are displayed in Figure~\ref{figpdfs_9_1}c and \ref{figpdfs_9_1}d, respectively. The true shear-wave velocity models of the synthetic cases are plotted in blue solid lines, the mean prediction models are plotted in green dashed lines and the MAP prediction models are plotted in red chain lines. The background color behind the lines denotes the 1D PPD at each layer. The shade of the color indicates the probability or uncertainty of the solution (darker colors represent higher probability). Note that the probability is normalized at each layer. 

In Figure~\ref{figpdfs_9_1}, the velocity models (Figure~\ref{figpdfs_9_1}c and \ref{figpdfs_9_1}d) are relatively simple. All the true solutions can be covered by the uncertainty of the solutions very well. In Figure~\ref{figpdfs_9_1}c, the MAP prediction model and mean prediction model fit the true model well at all layers, except for the last layer. The large uncertainty of the solution at this layer indicates the velocity is not very well determined for this layer. Relatively speaking, the mean prediction model gives a closer result to the true model than the MAP prediction model. Contrary to Figure~\ref{figpdfs_9_1}c, the MAP prediction model of Figure~\ref{figpdfs_9_1}d performs slightly better than the mean model. The overall uncertainty of the predictions in Figure~\ref{figpdfs_9_1}d is larger than Figure~\ref{figpdfs_9_1}c, which may be caused by the large uncertainty of the noisy dispersion data fed to the network.

The dispersion curves corresponding to the MAP and mean predictions are computed and displayed in Figure~\ref{figpdfs_9_1}a and \ref{figpdfs_9_1}b with red and green solid lines. The results show that although noisy data are fed to the network, the predictions can give comparable dispersion data to the true data. Furthermore, the slight misfit of the predictions in Figure~\ref{figpdfs_9_1}b indicates the impact of the noise added in the noisy data.

Figure~\ref{figpdfs_9_2} presents another two synthetic cases with more complicated velocity structures. Figure~\ref{figpdfs_9_2}c shows that the predicted models can not perfectly fit the true model, especially for shallow layers. It can be explained by two possible reasons. One reason is that shallow layers are very thin and have limited sensitivity to the loss function, which increases the difficulty of accurate prediction. The other reason is that the dispersion data in higher frequencies have more impact on the velocities of shallower layers, as is proved by the depth sensitivity analysis in \citet{wu2020shear}. Hence, large interference of the dispersion data in higher frequencies may result in large bias of the velocities in shallow layers. Similar phenomenon also exists in Figure~\ref{figpdfs_9_2}b and \ref{figpdfs_9_2}c. Large interferences of the noisy data also have an obvious impact on the MAP and mean dispersion data in Figure~\ref{figpdfs_9_2}b in the frequency band of 0.5 to 1.5 Hz. Overall, the MDN can give acceptable predictions in Figure~\ref{figpdfs_9_2}c and \ref{figpdfs_9_2}d, considering the interference of the noise and the complexity of the velocity structures.

Generally speaking, the predicted MAP and mean velocity models in the synthetic cases fit the true model well, except for some small anomalies caused by the large error of noisy data at some frequencies. Where noise interferes more in the noisy dispersion curve, the predicted dispersion curve is more likely to deviate from the true value. These synthetic cases indicate a good generalization performance of the designed MDN.

\subsection{Real data case}

\begin{figure} 
      \centering
      \includegraphics[width=3.5in, trim=0 10 0 20,clip]{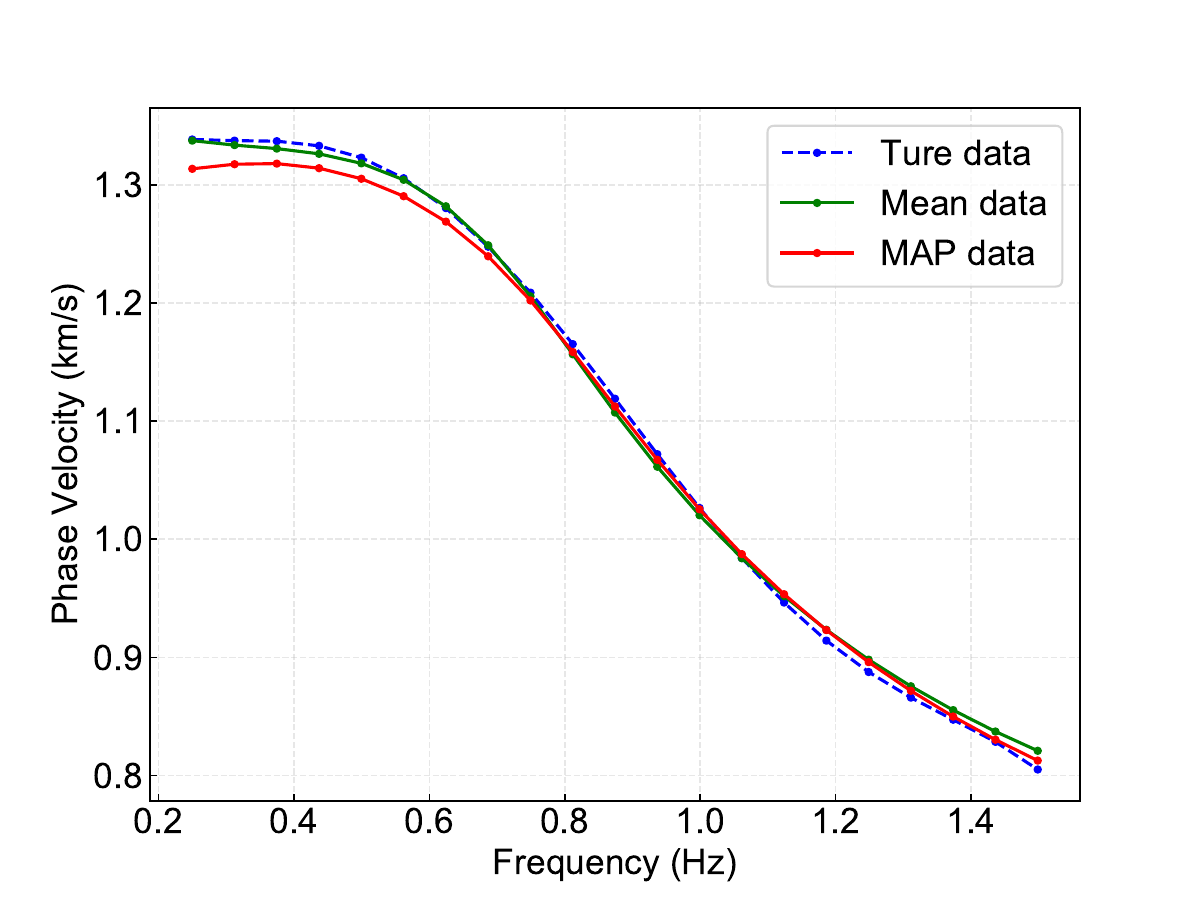}
      \caption{Dispersion curves extracted from marine ambient noise recorded at Snorre field by \citet{Li2012} and corresponding predicted dispersion data given by the MDN inversion.}
      \label{figdisps_snorre}
\end{figure}

\begin{figure} 
      \centering
      \includegraphics[width=\textwidth, trim=0 470 0 60,clip]{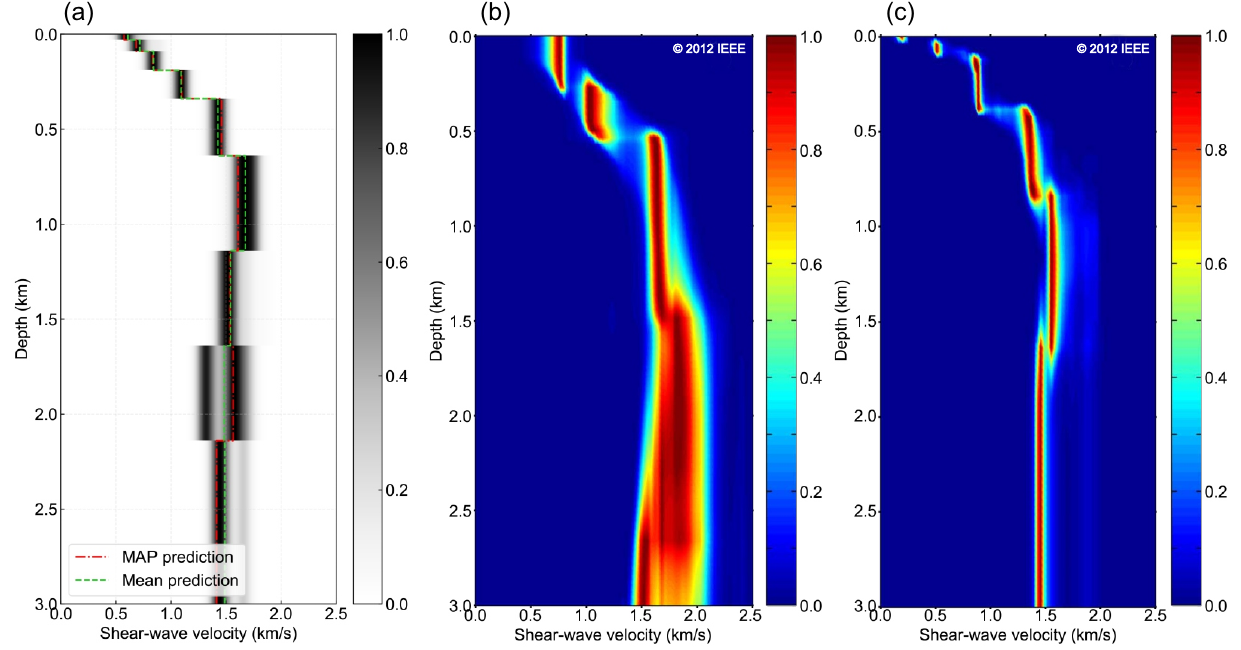}
      \caption{The performance of noisy training network compared with traditional Bayesian inversion results from \citet{Li2012}. (a) The prediction given by the MDN. (b) The prediction given by traditional Bayesian inversion using a 5-layer model. (c) The prediction given by traditional Bayesian inversion using a 8-layer model and 3 dispersion-curve modes.}
      \label{figstan_pred}
\end{figure}

Figure~\ref{figdisps_snorre} presents the dispersion curve (blue dashed line) extracted from ocean ambient noise recorded at Snorre field. Feed the data to the MDN and we can get the predicted velocity model shown in Figure~\ref{figstan_pred}a. For comparison purposes we present the traditional Bayesian inversion results given by \citet{Li2012} in Figure~\ref{figstan_pred}b and \ref{figstan_pred}c as well. Figure~\ref{figstan_pred}b shows the prediction given by the traditional Bayesian inversion using a 5-layer model and the dispersion curve shown in Figure~\ref{figdisps_snorre}. Figure~\ref{figstan_pred}c shows the prediction given by the traditional Bayesian inversion using a 8-layer model and 3 dispersion-curve modes (only the fundamental mode is shown in Figure~\ref{figdisps_snorre}, the other two higher order modes are not presented here).

Figure~\ref{figstan_pred}a shows several successively varying layers at the depth range of 0 to 1.0 km, which is different to Figure~\ref{figstan_pred}b. The difference is related to the parameterized models used by the two methods. However, the result in Figure~\ref{figstan_pred}a may be closer to the reality that the seabed hardness gradually increases and the shear wave velocity increases accordingly. In the stratigraphic interval with depth greater than 1.5 km, both methods present larger uncertainty.

There is a greater degree of similarity between the results in Figure~\ref{figstan_pred}a and \ref{figstan_pred}c, although only the fundamental mode of the dispersion curves is used for Figure~\ref{figstan_pred}a. In the depth interval of 0 to 1.0 km, both of them show the formation with continuously increasing shear wave velocity; in the depth interval of 0.8-1.2 km, both methods give a shear wave velocity larger than 1.5 km/s; in the depth interval greater than 1.6 km, the shear wave velocity of the formation is estimated to be around 1.5 km/s. There are also some differences between the two results. First, the first layer of Figure \ref{figstan_pred}c gives a lower initial shear wave velocity. The possible reason is that Figure \ref{figstan_pred}c uses three dispersion modes for inversion, among which the two additional higher-order modes belong to higher frequency information, providing higher resolution for shallow layers; second, in the depth interval greater than 1.5 km, the prediction in Figure~\ref{figstan_pred}a has greater uncertainty.

In general, the MDN inversion (Figure~\ref{figstan_pred}a) gives a reasonable and comparable result with traditional Bayesian inversion (Figure~\ref{figstan_pred}b and \ref{figstan_pred}c). Substituting the MAP and mean predictions back into the DISPER80 forward model, we can obtain the corresponding MAP and mean dispersion data, as is shown in Figure~\ref{figdisps_snorre}. The result indicates that the mean dispersion data are in good agreement with the original dispersion curve, while the MAP dispersion data have a certain degree of deviation from the original data in the frequency band of 0.2 to 0.6 Hz. However, the magnitude of the deviation between them is relatively small, which is lower than the magnitude of the noise intensity control factor $\epsilon = 0.05$ assumed in the network training process. It is within the acceptable error range.

\section{Conclusion}

We have presented a Bayesian geoacoustic inversion approach using the mixture density network (MDN), which can provide probabilistic solutions to nonlinear inverse problems. The derivation of the important geoacoustic statistics makes it convenient to train the network directly on the whole parameter space and get the multidimensional posterior probability density of model parameters directly. 

The evaluation on the test data set indicates a good generalization performance of the designed network trained using noisy data. The inversion results of synthetic cases show that the predictions of the MDN fit the true velocity model and true dispersion data well, although some local layers may be influenced by large error in the noisy data. In the real data case, the MDN inversion also provides comparable result with the traditional Bayesian inversion method. Once trained, the MDN can give a prediction within seconds, indicating a huge efficiency advantage over tradition inversion methods. It provides an promising approach for real-time inversion.

\begin{acknowledgments}
      We acknowledge Equinor for providing the field data used in this research. Hefeng Dong thanks the Research Council of Norway and the industry partners of the GAMES consortium at NTNU for financial support (grant no. 294404) and Guoli Wu thanks the National Natural Science Foundation of China for financial support(grant no. 41705078). 
\end{acknowledgments}

\newpage

\bibliographystyle{gji}  %
\bibliography{reference}

\begin{thebibliography}{38}
\expandafter\ifx\csname natexlab\endcsname\relax\def\natexlab#1{#1}\fi

\bibitem[Bensen et~al.(2009)Bensen, Ritzwoller, \& Yang]{Bensen2009}
Bensen, G.~D., Ritzwoller, M.~H., \& Yang, Y., 2009.
\newblock A 3-d shear velocity model of the crust and uppermost mantle beneath
  the united states from ambient seismic noise, {\bf 177}, 1177--1196.

\bibitem[Bishop(1994)]{bishop1994mixture}
Bishop, C.~M., 1994.
\newblock Mixture density networks.

\bibitem[Bodin \& Sambridge(2009)]{Bodin2009}
Bodin, T. \& Sambridge, M., 2009.
\newblock Seismic tomography with the reversible jump algorithm, {\it
  Geophysical Journal International\/}, {\bf 178}(3), 1411--1436.

\bibitem[Brando~Guillaumes(2017)]{brando2017mixture}
Brando~Guillaumes, A., 2017.
\newblock {\it Mixture density networks for distribution and uncertainty
  estimation\/}, Master's thesis, Universitat Polit{\`e}cnica de Catalunya.

\bibitem[Brocher(2005)]{Brocher2005}
Brocher, T.~M., 2005.
\newblock {Empirical Relations between Elastic Wavespeeds and Density in the
  Earth's Crust}, {\it Bulletin of the Seismological Society of America\/},
  {\bf 95}(6), 2081--2092.

\bibitem[Caiti et~al.(1994)Caiti, Akal, \& Stoll]{Caiti1994}
Caiti, A., Akal, T., \& Stoll, R.~D., 1994.
\newblock Estimation of shear wave velocity in shallow marine sediments, {\bf
  19}, 58--72.

\bibitem[Cao et~al.(2020)Cao, Earp, de~Ridder, Curtis, \& Galetti]{Cao2020}
Cao, R., Earp, S., de~Ridder, S. A.~L., Curtis, A., \& Galetti, E., 2020.
\newblock Near-real-time near-surface 3d seismic velocity and uncertainty
  models by wavefield gradiometry and neural network inversion of ambient
  seismic noise, {\it {GEOPHYSICS}\/}, {\bf 85}(1), KS13--KS27.

\bibitem[Castagna et~al.(1985)Castagna, Batzle, \& Eastwood]{Castagna1985}
Castagna, J.~P., Batzle, M.~L., \& Eastwood, R.~L., 1985.
\newblock Relationships between compressional‐wave and shear‐wave
  velocities in clastic silicate rocks, {\it Geophysics\/}, {\bf 50}, 571--581.

\bibitem[Collins et~al.(1992)Collins, Kuperman, \& Schmidt]{Collins1992}
Collins, M.~D., Kuperman, W.~A., \& Schmidt, H., 1992.
\newblock Nonlinear inversion for ocean‐bottom properties, {\bf 92},
  2770--2783.

\bibitem[Dong \& Dosso(2011)]{Dong2011}
Dong, H. \& Dosso, S.~E., 2011.
\newblock Bayesian inversion of interface-wave dispersion for seabed shear-wave
  speed profiles, {\it {IEEE} Journal of Oceanic Engineering\/}, {\bf 36}(1),
  1--11.

\bibitem[Dong et~al.(2010)Dong, Liu, Thompson, \& Morton]{hefengdong2010}
Dong, H., Liu, L., Thompson, M., \& Morton, A., 2010.
\newblock Estimation of seismic interface wave dispersion using ambient noise
  recorded by ocean bottom cable, in {\em ECUA 2010\/}.

\bibitem[Dosso \& Wilmut(2006)]{Dosso2006}
Dosso, S. \& Wilmut, M., 2006.
\newblock Estimating data uncertainty in matched-field geoacoustic inversion,
  {\it IEEE J. Ocean. Eng\/}, {\bf 31}, 470--479.

\bibitem[Dosso(2002)]{Dosso2002}
Dosso, S.~E., 2002.
\newblock Quantifying uncertainty in geoacoustic inversion. i. a fast gibbs
  sampler approach, {\bf 111}, 129--142.

\bibitem[Dosso \& Dettmer(2011)]{Dosso2011}
Dosso, S.~E. \& Dettmer, J., 2011.
\newblock Bayesian matched-field geoacoustic inversion, {\it Inverse
  Problems\/}, {\bf 27}(5), 055009.

\bibitem[Dosso et~al.(1993)Dosso, Yeremy, Ozard, \& Chapman]{Dosso1993}
Dosso, S.~E., Yeremy, M.~L., Ozard, J.~M., \& Chapman, N.~R., 1993.
\newblock Estimation of ocean-bottom properties by matched-field inversion of
  acoustic field data, {\bf 18}, 232--239.

\bibitem[Dosso et~al.(2001)Dosso, Wilmut, \& Lapinski]{Dosso2001}
Dosso, S.~E., Wilmut, M.~J., \& Lapinski, A.-L.~S., 2001.
\newblock An adaptive-hybrid algorithm for geoacoustic inversion, {\bf 26},
  324--336.

\bibitem[Earp et~al.(2019)Earp, Curtis, Zhang, \& Hansteen]{Earp2019}
Earp, S., Curtis, A., Zhang, X., \& Hansteen, F., 2019.
\newblock Probabilistic neural network tomography across grane field (north
  sea) from surface wave dispersion data, {\it arXiv preprint arXiv:1908.09588
  (Not peer reviewed)\/}.

\bibitem[Galetti et~al.(2017)Galetti, Curtis, Baptie, Jenkins, \&
  Nicolson]{Galetti2017}
Galetti, E., Curtis, A., Baptie, B., Jenkins, D., \& Nicolson, H., 2017.
\newblock Transdimensional love-wave tomography of the british isles and
  shear-velocity structure of the east irish sea basin from ambient-noise
  interferometry, {\it Geophysical Journal International\/}, {\bf 208}, 36--58.

\bibitem[Gerstoft(1994)]{Gerstoft1994}
Gerstoft, P., 1994.
\newblock Inversion of seismoacoustic data using genetic algorithms and a
  posteriori probability distributions, {\it The Journal of the Acoustical
  Society of America\/}, {\bf 95}(2), 770--782.

\bibitem[Gerstoft(1995)]{Gerstoft1995}
Gerstoft, P., 1995.
\newblock Inversion of acoustic data using a combination of genetic algorithms
  and the gauss–newton approach, {\bf 97}, 2181--2190.

\bibitem[Gerstoft \& Michalopoulou(2000)]{Gerstoft2000}
Gerstoft, P. \& Michalopoulou, Z.-H., 2000.
\newblock Global optimization in matched field inversion, in {\em 4th European
  Conference on Underwater Acoustics (Italian National Research Council),
  Rome\/}, pp. 27--32, Citeseer.

\bibitem[Heard et~al.(1998)Heard, Hannay, \& Carr]{Heard1998}
Heard, G.~J., Hannay, D., \& Carr, S., 1998.
\newblock Genetic algorithm inversion of the 1997 geoacoustic inversion
  workshop test case data, {\bf 06}, 61--71.

\bibitem[Li et~al.(2012)Li, Dosso, Dong, Yu, \& Liu]{Li2012}
Li, C., Dosso, S.~E., Dong, H., Yu, D., \& Liu, L., 2012.
\newblock Bayesian inversion of multimode interface-wave dispersion from
  ambient noise, {\it {IEEE} Journal of Oceanic Engineering\/}, {\bf 37}(3),
  407--416.

\bibitem[McLachlan \& Basford(1988)]{mclachlan1988mixture}
McLachlan, G.~J. \& Basford, K.~E., 1988.
\newblock Mixture models: Inference and applications to clustering, {\it
  Inference and Applications to Clustering\/}, {\bf 38}(2).

\bibitem[Meier et~al.(2007{\natexlab{a}})Meier, Curtis, \&
  Trampert]{Meier2007a}
Meier, U., Curtis, A., \& Trampert, J., 2007{\natexlab{a}}.
\newblock Global crustal thickness from neural network inversion of surface
  wave data, {\it Geophysical Journal International\/}, {\bf 169}(2), 706--722.

\bibitem[Meier et~al.(2007{\natexlab{b}})Meier, Curtis, \&
  Trampert]{Meier2007b}
Meier, U., Curtis, A., \& Trampert, J., 2007{\natexlab{b}}.
\newblock Fully nonlinear inversion of fundamental mode surface waves for a
  global crustal model, {\it Geophysical Research Letters\/}.

\bibitem[Mosegaard \& Tarantola(1995)]{Mosegaard1995}
Mosegaard, K. \& Tarantola, A., 1995.
\newblock Monte carlo sampling of solutions to inverse problems, {\it Journal
  of Geophysical Research: Solid Earth\/}, {\bf 100}(B7), 12431--12447.

\bibitem[Musil et~al.(1999)Musil, Wilmut, \& Chapman]{Musil1999}
Musil, M., Wilmut, M.~J., \& Chapman, N.~R., 1999.
\newblock A hybrid simplex genetic algorithm for estimating geoacoustic
  parameters using matched-field inversion, {\bf 24}, 358--369.

\bibitem[Saito(1988)]{Saito1980DISPER80}
Saito, M., 1988.
\newblock Disper80: A subroutine package for the calculation of seismic
  normal-mode solutions, in {\em Seismological Algorithms: Computational
  Methods and Computer Programs\/}, pp. 293--319, Academic Press, New York.

\bibitem[Shahraeeni et~al.(2012)Shahraeeni, Curtis, \& Chao]{Shahraeeni2012}
Shahraeeni, M.~S., Curtis, A., \& Chao, G., 2012.
\newblock Fast probabilistic petrophysical mapping of reservoirs from 3d
  seismic data, {\it {GEOPHYSICS}\/}, {\bf 77}(3), O1--O19.

\bibitem[Steininger et~al.(2014)Steininger, Dosso, Holland, \&
  Dettmer]{Steininger2014}
Steininger, G., Dosso, S.~E., Holland, C.~W., \& Dettmer, J., 2014.
\newblock A trans-dimensional polynomial-spline parameterization for
  gradient-based geoacoustic inversion, {\it The Journal of the Acoustical
  Society of America\/}, {\bf 136}(4), 1563--1573.

\bibitem[Tarantola \& Valette(1982)]{Tarantola1982}
Tarantola, A. \& Valette, B., 1982.
\newblock Generalized nonlinear inverse problems solved using the least squares
  criterion, {\bf 20}, 219.

\bibitem[Wit et~al.(2014)Wit, Käufl, Valentine, \& Trampert]{Wit2014}
Wit, R.~D., Käufl, P., Valentine, A., \& Trampert, J., 2014.
\newblock Bayesian inversion of free oscillations for earth's radial
  (an)elastic structure, {\it Physics of the Earth and Planetary Interiors\/},
  {\bf 237}, 1--17.

\bibitem[Wit et~al.(2013)Wit, Valentine, \& Trampert]{Wit2013a}
Wit, R. W. L.~D., Valentine, A.~P., \& Trampert, J., 2013.
\newblock Bayesian inference of earth's radial seismic structure from body-wave
  traveltimes using neural networks, {\bf 195}, 408--422.

\bibitem[Wu et~al.(2019)Wu, Dong, Ke, \& Song]{Wu2019}
Wu, G., Dong, H., Ke, G., \& Song, J., 2019.
\newblock An adapted eigenvalue-based filter for ocean ambient noise
  processing, {\it {GEOPHYSICS}\/}, {\bf 85}(1), KS29--KS38.

\bibitem[Wu et~al.(2020)Wu, Dong, Ke, \& Song]{wu2020shear}
Wu, G., Dong, H., Ke, G., \& Song, J., 2020.
\newblock Shear-wave tomography using ocean ambient noise with interference,
  {\it Remote Sensing\/}, {\bf 12}(18), 2969.

\bibitem[Zhang et~al.(2018)Zhang, Curtis, Galetti, \& de~Ridder]{Zhang2018a}
Zhang, X., Curtis, A., Galetti, E., \& de~Ridder, S., 2018.
\newblock 3-d monte carlo surface wave tomography, {\it Geophysical Journal
  International\/}, {\bf 215}(3), 1644--1658.

\bibitem[Zhdanov(2002)]{zhdanov2002}
Zhdanov, M.~S., 2002.
\newblock {\it Geophysical inverse theory and regularization problems\/},
  vol.~36, Elsevier.

\end{thebibliography}

\appendix
\section{}
\label{apif}
This section proves the following integral formula
\begin{equation}
      \int {\rm exp} \{ -\frac{1}{2\sigma_l^2} (m^\prime-\mu_{li})^2 \} {\rm d}m^\prime = \sqrt{2\pi}\sigma_l,
      \label{intmmu}
\end{equation}
where $\sigma_l$ and $\mu_l$ are considered to be constant. Note that $\int$ means integrating over the entire parameter space. For the convenience of calculation, we choose to compute the infinite integral as an estimation. Since the value interval of the parameter usually covers $[\mu_l-3\sigma_l,\mu_l+3\sigma_l]$ in geoacoustic inversions, the infinite integral would be a good estimation. First, we compute
\begin{equation}
      \iint_{-\infty}^{\infty} {\rm exp}\{-\left(x^{2}+y^{2}\right)\} {\rm d} x {\rm d} y=\int_{0}^{2 \pi} \int_{0}^{\infty} {\rm exp}\{-r^{2}\} r {\rm d} r {\rm d} \theta=\int_{0}^{2 \pi}-\left.\frac{1}{2} {\rm exp}\{-r^{2}\}\right|_{0} ^{\infty} {\rm d} \theta=\pi.
\end{equation}
On the other side, we have
\begin{equation}
      \iint_{-\infty}^{\infty} {\rm exp}\{-\left(x^{2}+y^{2}\right)\} {\rm d} x {\rm d} y=\int_{-\infty}^{\infty} {\rm exp}\{-x^{2}\} {\rm d} x \int_{-\infty}^{\infty} {\rm exp}\{-y^{2}\} {\rm d} y=\left(\int_{-\infty}^{\infty} {\rm exp}\{-x^{2}\} {\rm d} x\right)^{2}.
\end{equation}
It leads to the fact that
\begin{equation}
      \int_{-\infty}^{\infty} {\rm exp}\{-x^{2}\} {\rm d} x=\sqrt{\pi}.
      \label{intex2}
\end{equation}
Taking $y=m^\prime-\mu_{li}$, the left hand side (LHS) of Equation~\ref{intmmu} becomes
\begin{equation}
      {\rm LHS} = \int_{-\infty}^{\infty} {\rm exp}\{-\frac{1}{2\sigma_l^2} y^{2}\} {\rm d} y.
\end{equation}
Taking $z = y/(\sqrt{2}\sigma)$ and using Equation~\ref{intex2} we have
\begin{equation}
      \int_{-\infty}^{\infty} {\rm exp}\{-\frac{1}{2\sigma_l^2} y^{2}\} {\rm d} y = \int_{-\infty}^{\infty} {\rm exp}\{-z^{2}\} \sqrt{2} \sigma_l {\rm d} z = \sqrt{2\pi} \sigma_l.
\end{equation}

Another simple way to prove Equation~\ref{intmmu} is to use the conclusion that the integral of the standard normal distribution equals to 1
\begin{equation}
      \int_{-\infty}^{\infty} \frac{1}{\sqrt{2\pi}\sigma_l} {\rm exp} \{ -\frac{1}{2\sigma_l^2} (m^\prime-\mu_{li})^2 \} {\rm d}m^\prime = 1,
\end{equation}
which directly leads to the result of Equation~\ref{intmmu}.
\section{}
\label{apif1}

This section proves the following integral formula
\begin{equation}
      \int {(m_i-\overline{m}_i)^2}  {\rm exp} \{ -\frac{1}{2\sigma_l^2} (m_i -\mu_{li})^2 \} {\rm d}m_i  = \sqrt{2\pi}\sigma_l[(\overline{m}_i-\mu_{li})^2+\sigma_l^2],
      \label{apb1}
\end{equation}
where $\sigma_l$, $\mu_{li}$ and $\overline{m}_i$ are considered to be constant. The LHS of Equation~\ref{apb1} can be expanded to
\begin{equation}
      \begin{aligned}
            \frac{1}{\sqrt{2\pi}\sigma_l} {\rm LHS} 
            =& \int m_i^2 \frac{1}{\sqrt{2\pi}\sigma_l} {\rm exp} \{ -\frac{1}{2\sigma_l^2} (m_i -\mu_{li})^2 \} {\rm d}m_i - \\
            & \int 2\overline{m}_i m_i \frac{1}{\sqrt{2\pi}\sigma_l} {\rm exp} \{ -\frac{1}{2\sigma_l^2} (m_i -\mu_{li})^2 \} {\rm d}m_i + \\
            & \int \overline{m}_i^2 \frac{1}{\sqrt{2\pi}\sigma_l} {\rm exp} \{ -\frac{1}{2\sigma_l^2} (m_i -\mu_{li})^2 \} {\rm d}m_i \\
            =& I_1 - I_2 +I_3.
      \end{aligned}
      \label{apbi1i2i3}
\end{equation}
Using the integral property of standard normal distribution, we have
\begin{equation}
      \begin{aligned}
            \sigma_l^2
            =& \int (m_i-\mu_{li})^2 \frac{1}{\sqrt{2\pi}\sigma_l} {\rm exp} \{ -\frac{1}{2\sigma_l^2} (m_i -\mu_{li})^2 \} {\rm d}m_i  \\
            =& \int m_i^2 \frac{1}{\sqrt{2\pi}\sigma_l} {\rm exp} \{ -\frac{1}{2\sigma_l^2} (m_i -\mu_{li})^2 \} {\rm d}m_i - \\
            & \int 2\mu_{li}m_i \frac{1}{\sqrt{2\pi}\sigma_l} {\rm exp} \{ -\frac{1}{2\sigma_l^2} (m_i -\mu_{li})^2 \} {\rm d}m_i + \\
            & \int \mu_{li}^2 \frac{1}{\sqrt{2\pi}\sigma_l} {\rm exp} \{ -\frac{1}{2\sigma_l^2} (m_i -\mu_{li})^2 \} {\rm d}m_i \\
            =& I_1 - 2\mu_{li}^2 + \mu_{li}^2 \\
            =& I_1 - \mu_{li}^2,
      \end{aligned}
      \label{apbi1}
\end{equation}
\begin{equation}
      \begin{aligned}
            I_2 =& \int 2\overline{m}_i m_i \frac{1}{\sqrt{2\pi}\sigma_l} {\rm exp} \{ -\frac{1}{2\sigma_l^2} (m_i -\mu_{li})^2 \} {\rm d}m_i \\
            =& 2\overline{m}_i \mu_{li},
      \end{aligned}
      \label{apbi2}
\end{equation}
and
\begin{equation}
      \begin{aligned}
            I_3 =& \int \overline{m}_i^2 \frac{1}{\sqrt{2\pi}\sigma_l} {\rm exp} \{ -\frac{1}{2\sigma_l^2} (m_i -\mu_{li})^2 \} {\rm d}m_i \\
            =& m_i^2.
      \end{aligned}
      \label{apbi3}
\end{equation}
Substituting Equations~\ref{apbi1}, \ref{apbi2} and \ref{apbi3} back into Equation~\ref{apbi1i2i3} leads to
\begin{equation}
      {\rm LHS} = \sqrt{2\pi}\sigma_l[(\overline{m}_i-\mu_{li})^2+\sigma_l^2].
\end{equation}





\end{document}